\documentclass[runningheads]{llncs}

\usepackage[mobile]{eccv}

\usepackage[pagebackref=false,breaklinks=true,colorlinks,citecolor=eccvblue,bookmarks=true,bookmarksnumbered=true]{hyperref}
\hypersetup{
  pdftitle={Motion and Structure from Event-based Normal Flow},
  pdfsubject={Robotics, Computer Vision, State Estimation},
  pdfauthor={Zhongyang Ren, Bangyan Liao, Delei Kong, Jinghang Li, Peidong Liu, Laurent Kneip, Guillermo Gallego, Yi Zhou},
  pdfkeywords={Event Cameras, Motion Estimation, Asynchronous Sensor, High Dynamic Range, High Temporal Resolution, Normal optical flow}
}
\usepackage[absolute]{textpos}

\usepackage{eccvabbrv}

\usepackage{graphicx}
\usepackage{booktabs}

\usepackage{marvosym}

\usepackage{orcidlink}

\usepackage{multirow}

\usepackage{algorithm}
\usepackage{algorithmic}

\usepackage{hhline}
\usepackage{makecell}
\usepackage{diagbox}

\usepackage{cancel}

\def\bfc{\mathbf{c}}
\def\bfd{\mathbf{d}}
\def\bfe{\mathbf{e}}

\def\bfh{\mathbf{h}}

\def\bfj{\mathbf{j}}
\def\bfk{\mathbf{k}}

\def\bfn{\mathbf{n}}

\def\bfq{\mathbf{q}}

\def\bfs{\mathbf{s}}
\def\bft{\mathbf{t}}
\def\bfu{\mathbf{u}}

\def\bfx{\mathbf{x}}

\def\bfA{\mathbf{A}}
\def\bfB{\mathbf{B}}
\def\bfC{\mathbf{C}}
\def\bfD{\mathbf{D}}
\def\bfE{\mathbf{E}}

\def\bfH{\mathbf{H}}
\def\bfI{\mathbf{1}}

\def\bfM{\mathbf{M}}
\def\bfN{\mathbf{N}}
\def\bfO{\mathbf{O}}
\def\bfP{\mathbf{P}}

\def\ttB{\mathtt{B}}

\def\cX{\mathcal{X}}

\def\Rot{\mathtt{R}}

\def\bfomega{\boldsymbol{\omega}}

\def\bfnu{\boldsymbol{\nu}}

\newcommand{\os}[5]
{
{}^{#1}_{#2}{#3}^{#4}_{#5}
}

\makeatletter
\def\blfootnote{\xdef\@thefnmark{}\@footnotetext}
\makeatother

\def\angvel{\boldsymbol{\omega}}
\def\linvel{\boldsymbol{\nu}}
\def\bparams{\boldsymbol{\theta}}
\def\Real{\mathbb{R}}

\usepackage{adjustbox} %

\begin{document}

\title{Motion and Structure from Event-based\\ Normal Flow}

\definecolor{somegray}{gray}{0.5}
\newcommand{\darkgrayed}[1]{\textcolor{somegray}{#1}}
\begin{textblock}{11}(2.5, -0.1)  %
\begin{center}
\darkgrayed{This paper has been accepted for publication at the European Conference on Computer Vision (ECCV), 2024. \copyright Springer}
\end{center}
\end{textblock}

\titlerunning{Motion and Structure from Event-based Normal Flow}
\authorrunning{Z. Ren et al.}

\author{Zhongyang Ren\inst{1}\textsuperscript{*} \orcidlink{0009-0005-5095-3226} \and
Bangyan Liao\inst{2}\textsuperscript{*}\orcidlink{0009-0007-7739-4879} \and
Delei Kong\inst{1}\orcidlink{0000-0002-5681-587X} \and
Jinghang Li\inst{1}\orcidlink{0000-0001-6196-6165} \and \\
Peidong Liu\inst{2}\orcidlink{0000-0002-9767-6220} \and
Laurent Kneip\inst{3}\orcidlink{0000-0001-6727-6608} \and
Guillermo Gallego\inst{4}\orcidlink{0000-0002-2672-9241} \and
Yi Zhou\inst{1}\textsuperscript{\text{\Letter}}\orcidlink{0000-0003-3201-8873}}

\authorrunning{Z. Ren et al.}

\institute{School of Robotics, Hunan University \and
School of Engineering, Westlake University \and 
School of Information Science and Technology, ShanghaiTech University \and
TU Berlin, ECDF, SCIoI Excellence Cluster and Robotics Institute Germany
}

\maketitle
\begin{abstract}
Recovering the camera motion and scene geometry from visual data is a fundamental problem in computer vision.
Its success in conventional (frame-based) vision is attributed to the maturity of feature extraction, data association and multi-view geometry.
The emergence of asynchronous (event-based) cameras calls for new approaches that use raw event data as input to solve this fundamental problem.
State-of-the-art solutions typically infer data association implicitly by iteratively reversing the event data generation process. 
However, the nonlinear nature of these methods limits their applicability in real-time tasks, and the constant-motion assumption leads to unstable results under agile motion.
To this end, we reformulate the problem in a way that aligns better with the differential working principle of event cameras.
We show that event-based normal flow can be used, via the proposed geometric error term, as an alternative to the full (optical) flow in solving a family of geometric problems that involve instantaneous first-order kinematics and scene geometry. 
Furthermore, we develop a fast linear solver and a continuous-time nonlinear solver on top of the proposed geometric error term.
Experiments on both synthetic and real data show the superiority of our linear solver in terms of accuracy and efficiency, and its practicality as an initializer for previous nonlinear solvers.
Besides, our continuous-time non-linear solver exhibits exceptional capabilities in accommodating sudden variations in motion since it does not rely on the constant-motion assumption. 
Our project page can be found at 
\href{https://nail-hnu.github.io/EvLinearSolver/}{https://nail-hnu.github.io/EvLinearSolver/}.


\end{abstract}

\blfootnote{$\ast$ equal contribution; $\text{\Letter}$ corresponding author (eeyzhou@hnu.edu.cn).}
\section{Introduction}
\label{sec:introduction}
Neuromorphic event-based cameras are bio-inspired sensors that report only brightness changes at each pixel location asynchronously.
Their high temporal resolution enables them to sense at exactly the same rate as the scene dynamics, and thus, event cameras are suitable for dealing with robotic perception~\cite{Rebecq16bmvc,Zhou18eccv,Zhu18eccv,Stoffregen19iccv,Shiba22eccv}, navigation~\cite{Kim16eccv,Rebecq17ral,Rebecq17bmvc,zhou2021esvo}, and control~\cite{Conradt09iscas,Delbruck13fns} tasks that involve agile motion.
A general and fundamental problem among these tasks is to recover the camera's ego-motion and scene geometry from event data.
One of the bottlenecks for solving this kind of problems at very high speeds is the lack of methods for efficient and robust data association.

Unlike its standard-vision counterparts, an event camera cannot observe visual information (\ie~intensity changes) when motion is parallel to edges.
Such a partial observability of event data makes optical flow estimation (as well as feature detection and matching) an ill-posed problem.
To this end, many existing methods \cite{Gallego18cvpr, nunes2020entropy, liu2021spatiotemporal, Gu21iccv, huang2023progressive} for solving the problem of motion-model fitting on event data propose specific metrics that assess event-data association established in either an implicit manner or an explicitly model-driven style.
The best fit can be retrieved by iteratively optimizing the data association, often expressed in terms of event alignment.
However, we argue two flaws of these methods: 
($i$) The success of these non-linear solutions requires a proper initialization.
While a brute-force search may be applicable, it is computationally expensive for real-time application, specially if the dimension of the parameter is high.
Efficient and deterministic initialization methods are lacking and urgently needed.
($ii$) The common assumption of constant velocity only holds within a short time interval.
Nevertheless, a time window that is too narrow may contain insufficient number of events for a valid model fitting.
The heuristic of using a constant number of events may violate the constant velocity assumption, especially in the case of sudden changes in speed.

\begin{figure}[t]
\centering
\includegraphics[width=0.95\textwidth]{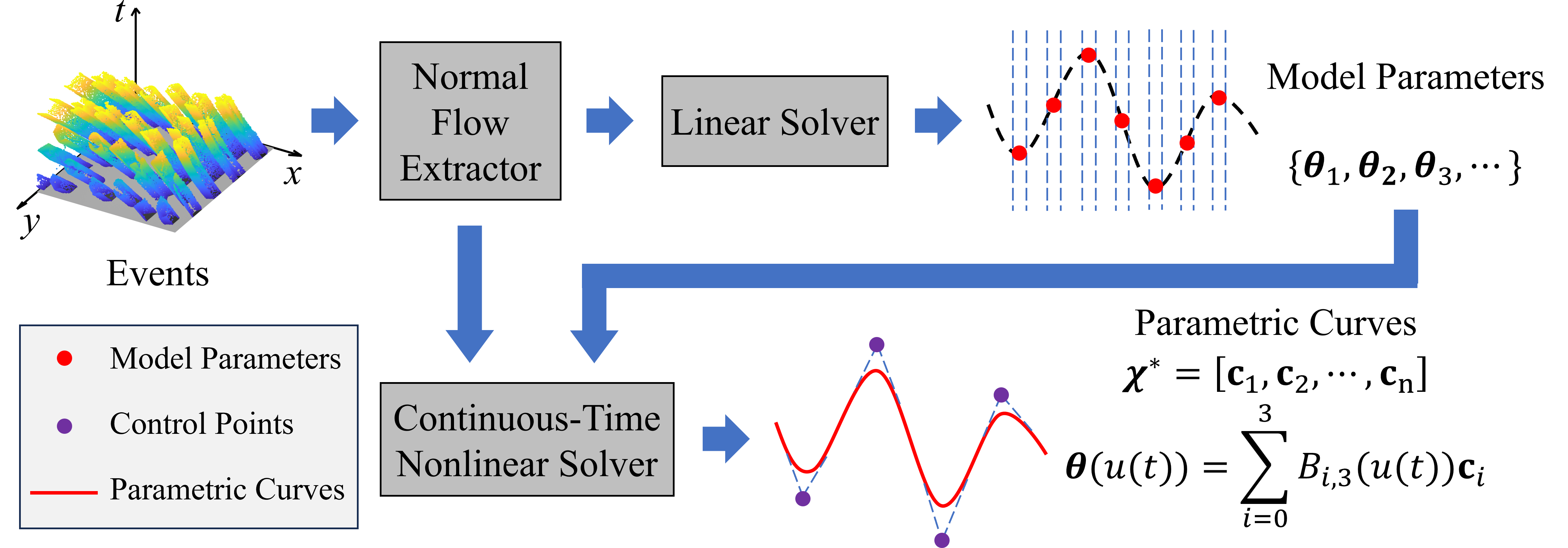}
\caption{\emph{Overview of our event-based motion estimation approach}.
Normal flow vectors are computed from the raw event data and are used as input to the two solvers.
The result of the linear solver consists of fitted motion models $\{\boldsymbol{\theta}_1, \boldsymbol{\theta}_2, \cdots \}$ at discrete time instants, which can be used to initialize the continuous-time nonlinear solver.}
\label{fig:our_slovers}
\end{figure}

In this paper, we rethink the problem formulation in a way that is more consistent with the differential working principle of event cameras. 
Unlike existing pipelines that use events occurring within a time interval to resolve the relative motion, we attempt to recover the structure and instantaneous first-order kinematic parameters involved in the motion field equation, using as input an instantaneous observation, namely event-based normal flow.
To this end, we explore effective constraints between normal-flow observations and model parameters for solving a family of motion-and-structure estimation problems, such as the estimation of depth and full optical flow, instantaneous angular velocity and linear velocity, and differential homography (see \cref{fig:our_slovers}).

The contributions of this paper can be summarized as follows:
\begin{itemize}
    \item A normal flow constraint with a geometric connection to event data.
    It overcomes the partial observability issue (\ie, errors caused by using normal flow as replacement for full optical flow) in general problems of geometric model fitting when normal flow is used as input.
    
    \item Two solvers for a family of motion-and-structure estimation problems using as input sparse event-based normal flow.
    Our linear solver (can be used with RANSAC) leads to closed-form and deterministic solutions that can be used as an initialization to existing nonlinear methods.
    Our nonlinear solver, under a continuous-time formulation, can handle sudden speed variations, and thus, is free of the constant-motion assumption.
    
    \item A thorough evaluation on the two proposed solvers, comparing against state-of-the-art methods, and also including an investigation on the performance of existing non-linear methods initialized using our linear solver.
\end{itemize}

\emph{Outline}: The rest of the paper is organized as follows.
We first discuss the related work in Sec.~\ref{sec:related_work}.
Then a compact review of preliminary knowledge is given in Sec.~\ref{sec:preliminaries}.
Our approach is presented in Sec.~\ref{sec:methodology}, including the normal-flow constraint, the linear models of a family of motion-and-structure estimation problems, and the proposed linear and nonlinear solvers.
Experimental evaluation is conducted in Sec.~\ref{sec:evaluation} and conclusion are drawn in Sec.~\ref{sec:conclusions}.

\section{Related Work}
\label{sec:related_work}

\subsection{Event-based Geometric Model Fitting}
\label{subsec:event_based_geometric_model_fitting}

The major body of literature on this specific topic follows an iterative scheme, which typically has its own special metric evaluating quantitatively the association between events, and solve the model fitting as a nonlinear optimization problem.
Among the first works is the \emph{Contrast Maximization} method \cite{Gallego18cvpr,Gallego19cvpr}, which provides a unified framework for solving a family of motion-model fitting problems.
It searches the optimal model parameters that ultimately lead to a sharp image of warped events (IWE).
The goodness of fit is evaluated by the alignment of all involved events along point trajectories on the image plane, and such alignment is assessed by the strength of the contours of the IWE.
The computational complexity of CMax is high because of iteratively updating of IWE, whose cost scales with the spatial resolution and the amount of event data.

To lessen the dependency on IWE, Nunes \etal proposes an entropy minimization framework (EMin) \cite{nunes2020entropy} that replaces the dispersion metric of \cite{Gallego18cvpr} with a similarity measurement between events, enabling to evaluate the goodness of fit in a higher dimensional space.
EMin is, still, too expensive for real-time application.
Liu \etal \cite{liu2021spatiotemporal} propose a two-step iterative method for the problem of 3-D rotation estimation using events as input.
Given a rotation estimate, the first step determines event correspondences by looking for the nearest neighbour in the spatio-temporal domain.
The second step employs a spatio-temporal registration which aligns corresponding events by minimizing the sum of squared geodesic residual.
The least-square nature of the energy function in the second step leads to a computation friendly performance.
Gu \etal \cite{Gu21iccv} present a probabilistic approach for spatio-temporal alignment of events.
The aligned events at a particular pixel are modeled as a spatio-temporal Poisson point process and the goodness of fit can be quantified using the likelihood of the data.
The optimal model is retrieved when the likelihood is maximized.
More recently, an event-to-map alignment scheme is proposed in \cite{huang2023progressive}, which adopts the time surface (TS) map as a reference for event alignment.
The proposed TS loss function drives later events to get increasingly closer to pixel positions of corresponding earlier events, thus, leading to a progressively sharp TS map as iterations proceed. 

In summary, all the above-mentioned methods are nonlinear, and thus, a proper initialization is necessary.
Although a brute-force search is widely adopted, sometimes it can be too expensive to locate the convergence basin in parameter space.
Moreover, \cite{Gallego18cvpr,nunes2020entropy} may come across a convergence to false-positive minima, known as event collapse \cite{Shiba22sensors,Shiba22aisy}.
Hence, a quick and deterministic solver (e.g., a closed-form solution) is desirable.
Besides, most of these methods assume a constant-motion model, which limits their applicability to scenarios of sudden speed variations.
In addition, we have noticed that a recent work \cite{gao20235} can jointly estimate the line parameters and the camera's linear velocities. 
However, this method is not a general one due to the dependence on line structures and the usage of an IMU.
Thus, it is not within the scope of our discussion.

\subsection{Motion and Structure from Normal Flow}
\label{subsec:motion_and_structure_from_normal_flow}

Due to the aperture problem, a local estimate of optical flow recovers only the partial component of the full flow vector along the direction of image gradient.
Preliminary knowledge about normal flow can be found in the supplementary material.
Recovering motion and structure from normal flow on intensity images is a well-studied topic.
Its main body of literature typically proposes a qualitatively deterministic solution by which a set (even infinite) of applicable ego motions are determined via an analysis of normal flow classification \cite{fermuller1995passive,brodsky2000structure,ji20063d}.
Only recently, researches devised ways to formulate the Cheirality constraint as a differentiable term and use it in a continuous optimization approach \cite{barranco2021joint} or as a loss function for training an artificial neural network \cite{parameshwara2022diffposenet}.

In event-based perception and navigation, similar methodologies as the frame-based ones are also witnessed, where normal flow computed from event data is used as input to determine the Focus of Expansion (FoE) for drones' course estimation \cite{dinaux2021faith} and obstacle avoidance \cite{Clady14fns}.

Most of these heuristic methods, while effective, are cumbersome for practical application.
In contrast, our method delivers a deterministic way to retrieve motion and structure from normal flow.
Additionally, it is noticed that normal flow can replace full flow in some specific problems of geometric model fitting, such as affine models \cite{meyer1992estimation, meyer1994time}.
In this way, the geometric model fitting problem can be easily solved in closed-form.
Nevertheless, these methods do not consider the observability bias in normal flows, and thus, such naive substitution may lead to inaccurate model fitting.
To this end, we present a robust linear solver based on a carefully-designed geometric measurement model that can overcome the partial observability of event-based normal flow.
{It is worth mentioning that a recent work \cite{lu2024eviv} also determines instantaneous linear velocity using as input the normal flow.
Our difference is seen in the scope of applicability. 
The usage of normal flow is restricted in the motion flow equation of \cite{lu2024eviv}, while our paper extends it to a wider scope (\ie, a family of geometric model fitting problems) by combining normal flow with \eg, differential epipolar geometry, differential homography, \etc.}

\section{Preliminaries}
\label{sec:preliminaries}
To make this paper self-contained, some preliminary knowledge is revisited.
Due to space limitations, here we only list essential concepts. 
A more detailed review of preliminary knowledge can be found in the supplementary material.

\subsubsection{Motion Field.}
Consider a perspective camera moving in a static environment, with instantaneous angular and linear velocities in the camera frame $\os{}{}\bfomega{}{} = (\omega_{x},\omega_{y},\omega_{z})^{\top}$ and $\os{}{}\bfnu{}{} = (\nu_{x},\nu_{y},\nu_{z})^{\top}$, respectively.
The motion field at pixel $\bfx = (x,y)^\top$ produced by projecting the 3D motion at point with depth $Z(\bfx)$ is
\begin{equation}
\label{eq:motion_flow_equation}
\bfu = \frac1{Z(\bfx)} \mathbf{A}(\bfx) \os{}{}\bfnu{}{} + \mathbf{B}(\bfx) \os{}{}\bfomega{}{},
\end{equation}
where 
$\mathbf{A}(\bfx)$ and $\mathbf{B}(\bfx)$ are $2\times 3$ matrices that solely depend on $\bfx$.
Equation \eqref{eq:motion_flow_equation} is the motion field equation \cite[Ch.8]{Trucco98book}\cite{longuet1980interpretation}.
Optical flow in rigid scenes aims at estimating the motion field from visual data, which is only possible in regions that do not suffer from the aperture problem.

\subsubsection{Brightness constancy} is the assumption stating that the true motion trajectories $\bfx(t)$ followed by objects on the image plane are those at which brightness is constant: $I(\bfx(t),t)=\text{const}$. 
Differentiating in time (total derivative) gives the equation at $(\bfx,t)$: 
$\nabla I\cdot \dot{\bfx} + \partial_t I = 0$,
where $\dot{\bfx} = d\bfx/dt$ is the velocity of image point $\bfx(t)$. 
Given $I$, this equation provides one constraint per point $(\bfx,t)$ to estimate $\dot{\bfx}$, now called optical flow. 
Without additional equations, only the normal component of the optical flow can be determined \cite{Trucco98book},
\begin{equation}
\bfn 
= - \frac{\partial_t I}{\|\nabla I\|^2} \nabla I.
\end{equation}

\subsubsection{Normal Flow.} 
Decomposing the optical flow $\bfu = \bfu_{\perp} + \bfu_{\parallel}$ into its normal and parallel components to the local image gradient ($\nabla I$), 
the normal flow is $\bfn = \bfu_{\perp}$, 
and the dot-product with $\bfn$ gives
\begin{equation}
\label{eq:normalflow:dotprod}
\bfn^\top \bfu  = \|\bfn\|^2, \quad\text{ or equivalently }\quad (\bfu- \bfn)\cdot \bfn = 0.
\end{equation}

\subsubsection{Differential Homography.}
(\emph{Continuous Homography Matrix} \cite[Ch.5]{Ma04book}).
If all scene points lie on a plane, their coordinates are related to their image projections via a homography transformation (which encodes the information of the scene plane $\{\bfN, d\}$ and the camera pose). 
In this case, the motion field \eqref{eq:motion_flow_equation} can be rewritten to adopt the form in homogeneous coordinates 
\begin{equation}
    \label{eq:differ_homo}
    \hat{\bfu}(\bfx) = \left(\bfI - \hat{\bfx}\bfe_3^{\top}\right)\bfH_d\hat{\bfx},
\end{equation}
where $\bfH_{d} \doteq -\left([\bfomega]_{\times} + d^{-1} \bfnu\bfN^{\top}\right)$ denotes the differential homography matrix ($[\cdot]_\times$ refers to the cross-product, skew-symmetric matrix), 
$\bfI$ is the identity matrix, 
$\hat{\bfx} = (x,y,1)^\top$, $\hat{\bfu} = (u_x, u_y, 0)^\top$ and $\bfe_3 = (0,0,1)^{\top}$.

\subsubsection{Differential Epipolar Geometry.}
(\emph{Continuous Epipolar Geometry} \cite[Ch.5]{Ma04book}).
The epipolar geometry examines the distance between a potential match to the determined epipolar line, and can be expressed as $\hat{\bfx}^{\prime\top}\bfE\hat{\bfx} = 0.$
The differential epipolar geometry can be derived analogously,
providing a relationship between $\bfx,\bfu$ and the camera velocities $\linvel,\angvel$.
In homogeneous coordinates:
\begin{equation}
    \label{eq:diff_epi}
    \hat{\bfu}(\bfx)^{\top}[\linvel]_\times \hat{\bfx} - \hat{\bfx}^{\top} \bfs \hat{\bfx} = 0,
\end{equation}
where $\bfs = \frac{1}{2}([\linvel]_\times[\angvel]_\times+[\angvel]_\times [\linvel]_\times)$ is a $3\times 3$ symmetric matrix.

\section{Methodology}
\label{sec:methodology}

\Cref{fig:our_slovers} summarizes our approach, grounded upon event-based normal flow. 
In this section, we first disclose the normal-flow constraint that can overcome the partial observability of normal flow when replacing full optical flow with normal flow in general tasks of geometric model fitting (\cref{subsec:normal flow constraint}).
By combining the constraint and the theory in \cref{sec:preliminaries}, we derive linear equations for a family of motion-and-structure estimation problems (\cref{subsec:models of motion ans structure}).
To solve these problems, we present a linear solver and a continuous-time nonlinear solver (\cref{subsec:our solvers}).

\subsection{Normal Flow Constraint}
\label{subsec:normal flow constraint}

Normal flow can be computed easily from the event data, for example by fitting planes locally to the space-time ``cloud'' of events on image space \cite{Benosman14tnnls,pijnacker2018vertical}.
Specifically, assuming that the event data $e_k \doteq \{\bfx_{k}, t_{k}\}$ spans locally a surface in the spatio-temporal domain,
the surface can be defined by a function $\mathrm{\Sigma}_{e}: \Real^2 \to \Real; \bfx_k \mapsto t_k$ that maps the pixel coordinates to the latest event's timestamp \cite{Benosman14tnnls,Clady14fns}. 
The normal flow is parallel to the spatial derivative of $\mathrm{\Sigma}_{e}$, 
namely ${\nabla}\mathrm{\Sigma}_e(\bfx)$.
More specifically, it can be calculated by 
\begin{equation}
    \label{eq:normal_flow_computtion_from_events}
    \bfn(\bfx) = \frac{{\nabla}\mathrm{\Sigma}_e(\bfx)}{\left \|  {\nabla }\mathrm{\Sigma}_e(\bfx)\right \|^2}.
\end{equation}
Readers may refer to the supplementary material for a more detailed derivation.
Surely there are other methods to extract event-based normal flow 
and the solvers we develop in this section are independent of them, 
i.e., our solvers can also be fed with the extracted normal flow data from such methods.

Event-based normal flow cannot be used directly as a replacement for the full flow in \eqref{eq:motion_flow_equation} to solve for the motion parameters.
However, we can leverage the relationship between both flows \eqref{eq:normalflow:dotprod} 
to define an effective geometric error criterion that circumvents the partial observability of normal flow.
If the flow $\bfu$ depends on some unknown (``state'') variables $\bparams$, we propose estimating them based on how well the normal flow constraint \eqref{eq:normalflow:dotprod} is satisfied, 
e.g., by minimizing some function of the error terms  %
\begin{equation}
    \label{eq:geometric_error}
     \bfn(\bfx)^\top \bfu(\bfx; \bparams) - \|\bfn(\bfx)\|^2 \doteq \epsilon_{\text{nf}} \,\in \Real.
\end{equation}

\begin{figure}[t]
\centering
\begin{subfigure}{0.213\linewidth}
\includegraphics[width=\textwidth]{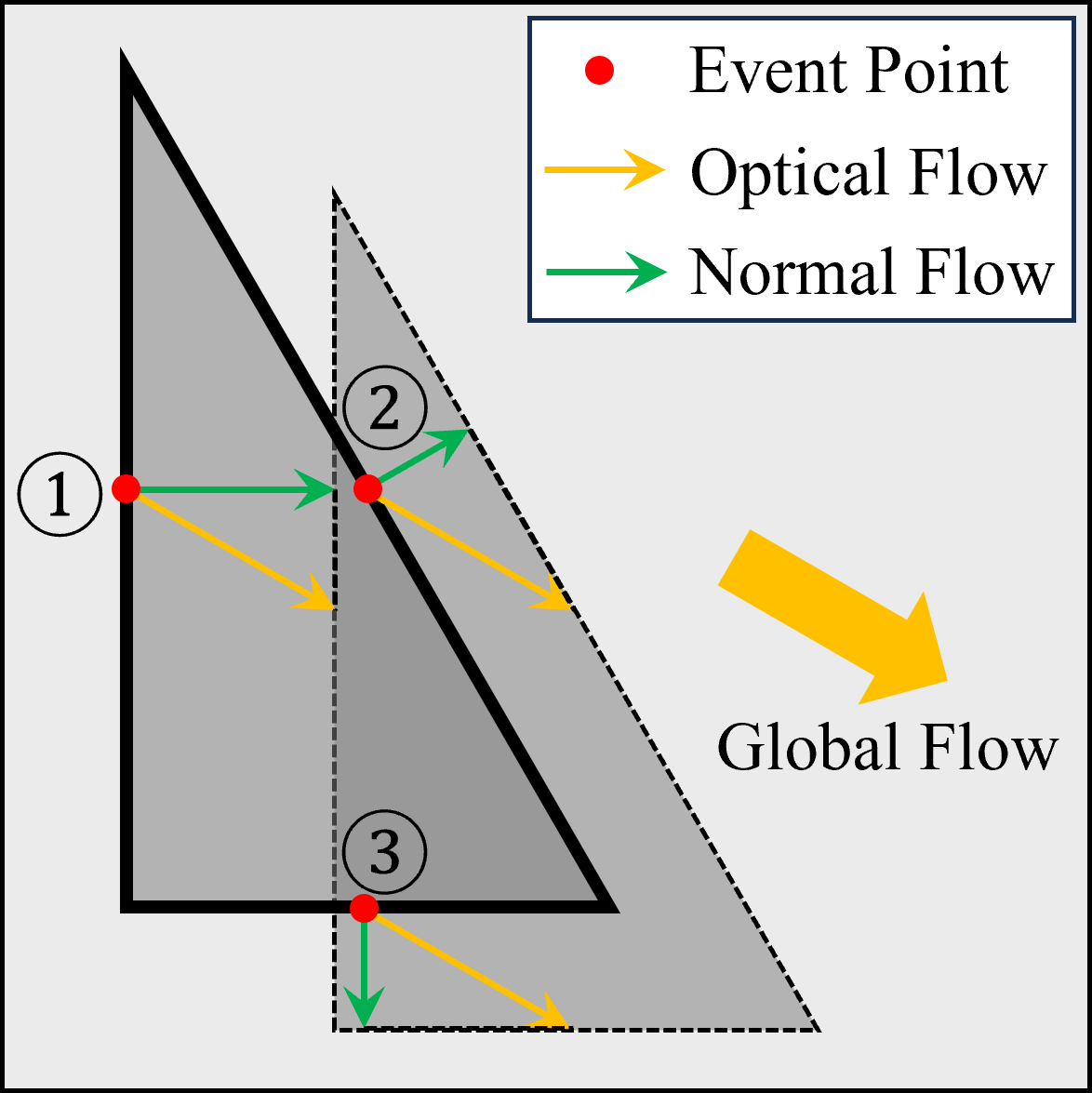}
\caption{}
\label{fig:2D-2D registration task}
\end{subfigure}
\begin{subfigure}{0.253\linewidth}
\includegraphics[width=\textwidth]{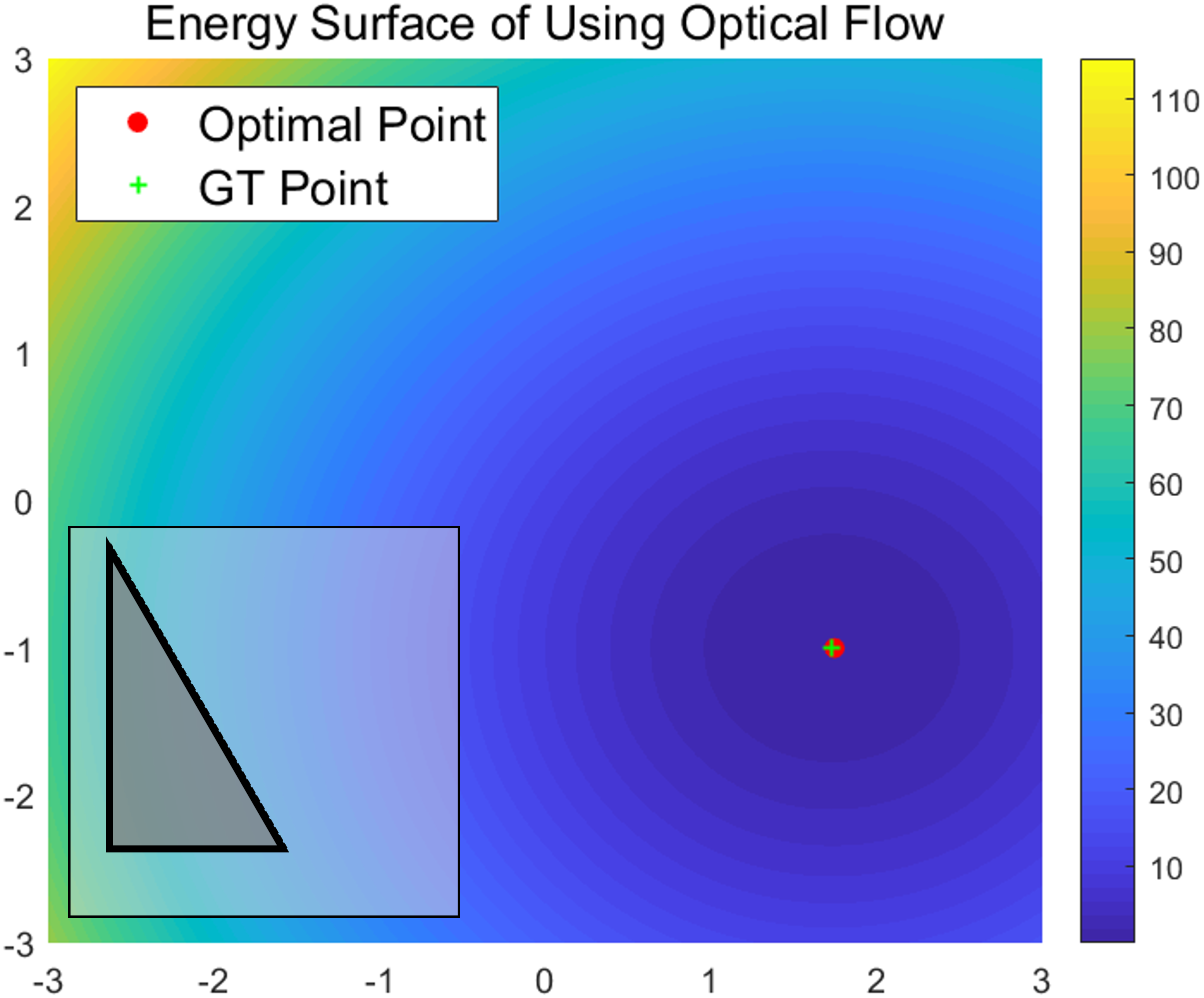}
\caption{}
\label{fig:Energy surface of using optical flow}
\end{subfigure}
\begin{subfigure}{0.253\linewidth}
\includegraphics[width=\textwidth]{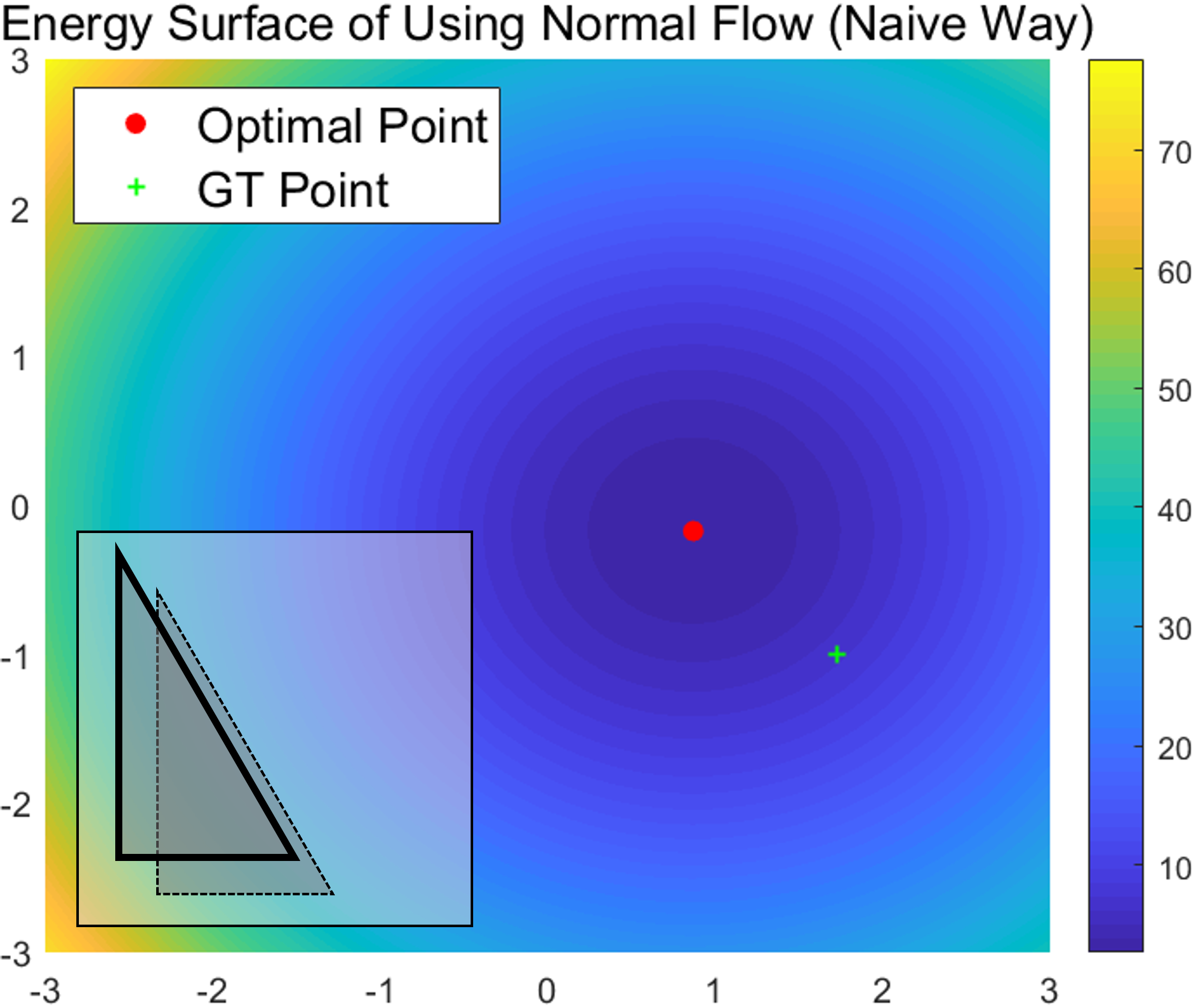}
\caption{}
\label{fig:Energy surface of using normal flow naively}
\end{subfigure}
\begin{subfigure}{0.253\linewidth}
\includegraphics[width=\textwidth]{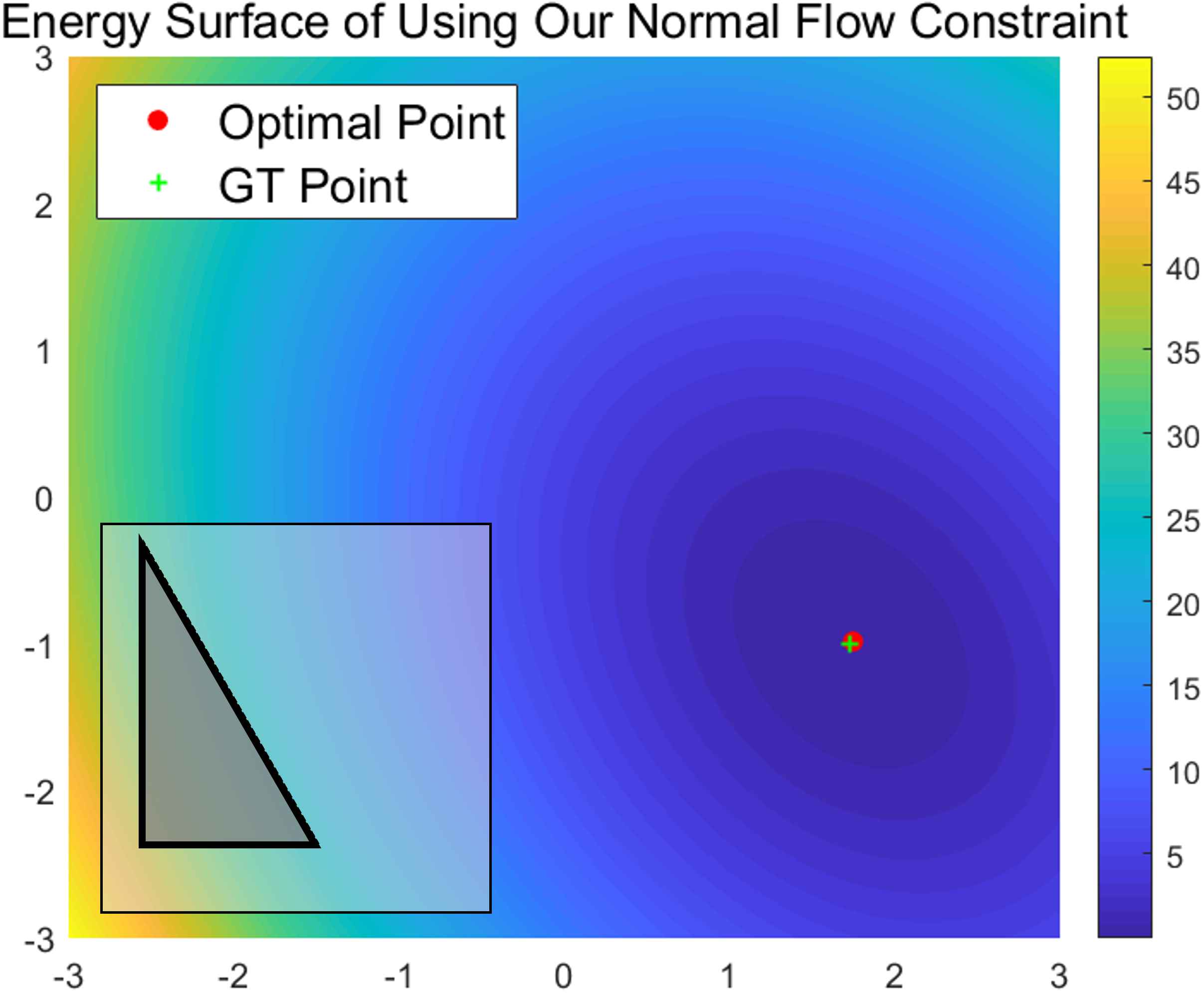}
\caption{}
\label{fig:Energy surface of using normal flow constraint}
\end{subfigure}
\caption{Toy example of applying the proposed normal-flow constraint.
(a): A 2D registration task using as input either optical flow or normal flow. 
The groundtruth displacement is defined by a global flow of $(1.732, -1)^\top$.
(b)-(d): The loss landscape obtained using different geometric measurements in the registration task. 
The red dot denotes the resulting displacement, and the green cross the groundtruth displacement.
The bottom-left regions in (b)-(d) display the corresponding registration results.}
\label{fig:toy_example}
\end{figure}

To illustrate the validity of the proposed normal-flow constraint over the naive way of simply using normal flow to replace full optical flow, we use a 2D registration task as toy example.
As shown in \cref{fig:toy_example}, the task consists of registering the solid-line triangle to the dashed-line one using as input either optical flow or normal flow.
We observe that using the proposed normal-flow constraint as a geometric measurement leads to a registration result that is almost as accurate as the one given by the full optical flow.
However, due to the partial observability of normal flow, simply using it as a replacement for full optical flow leads to a biased registration result, as shown in \cref{fig:Energy surface of using normal flow naively}.
As we see in all following experiments, similar conclusions can be made in general problems of geometric model fitting.

\subsection{Models of Motion and Structure}
\label{subsec:models of motion ans structure}
Based on the derivation in \cref{subsec:normal flow constraint}, we establish linear models to a family of motion and structure estimation problems using the event-based normal flow as input.
In particular, we first discuss how to recover optical flow and depth by introducing the normal-flow constraint into the differential epipolar geometry.
Then we show that the combination of the normal-flow constraint with the motion flow equation \eqref{eq:motion_flow_equation} results in solutions to the problem of instantaneous angular velocity estimation, and 6-DoF motion estimation given depth as prior, respectively.
Finally, we show how to compute the differential homography.

\subsubsection{Optical Flow.}
\label{subsubsec:optical_flow_and_depth_estimation}
Given as input the accurate normal flow, and camera velocities, the full optical flow $\bfu(\bfx)$ at pixel location $\bfx$ can be calculated by solving the following linear system obtained from the combination of the normal-flow constraint \eqref{eq:normalflow:dotprod} and the differential epipolar geometry \eqref{eq:diff_epi}: 
\begin{equation}
    \label{eq:optical_flow_estimation}
    \begin{pmatrix}
        {\bfn}(\bfx)^{\top}\\[0.5ex]
        -\hat{\bfx}^{\top}[\linvel]_\times 
    \end{pmatrix}
    {\bfu}(\bfx) = 
    \begin{pmatrix}
        \|\bfn(\bfx)\|^2 \\[0.5ex]
        \hat{\bfx}^{\top} \bfs \hat{\bfx}
    \end{pmatrix}.
\end{equation}

\subsubsection{Depth Estimation.}
Similarly, if the normal flow and camera velocities are known, 
the scene depth $Z(\bfx)$ can be determined by 
combining the normal-flow constraint \eqref{eq:normalflow:dotprod} with the motion field eq. \eqref{eq:motion_flow_equation}, 
yielding the closed-form solution:
\begin{equation}
\label{eq:depth inference}
    Z(\bfx) = \frac{\bfn(\bfx)^{\top}\bfA(\bfx)\linvel}
    {\|\bfn(\bfx)\|^2 - {\bfn}(\bfx)^{\top}\bfB(\bfx)\angvel}.
\end{equation}

\subsubsection{Angular Velocity Estimation.}
\label{subsubsec:angular_velocity_estimation}
Assuming that an event camera undergoes a pure rotational motion, the optical flow $\bfu$ in \eqref{eq:motion_flow_equation} can be simplified as
\begin{equation}
    \label{eq:rotation_flow}
    \bfu(\bfx) = \bfB(\bfx)\angvel.
\end{equation}
By substituting \eqref{eq:rotation_flow} into \eqref{eq:normalflow:dotprod}, we have the linear equation in $\angvel$:
\begin{equation}
\label{eq:rotation_single}
\bfn(\bfx)^{\top} \bfB(\bfx) \angvel = \|\bfn(\bfx)\|^2. 
\end{equation}
Since each normal flow provides only one constraint, the minimal problem requires three normal flow vectors as input.

\subsubsection{6-DoF Motion Estimation.}
\label{subsubsec:6_dof_motion_estimation_with_depth_priors}
Assume the event camera undergoes a general 6-DoF motion with known depth of the observed scene,
we can solve the first-order kinematics, namely instantaneous linear and angular velocities, by solving the following linear system obtained from the normal-flow constraint \eqref{eq:normalflow:dotprod} and the motion-flow equation \eqref{eq:motion_flow_equation}:
\begin{equation}
    \label{eq:absolute_pose}
    \bfn(\bfx)^{\top}
    \bfD(\bfx)
    \begin{bmatrix}
    \linvel\\
    \angvel
    \end{bmatrix} = \|\bfn(\bfx)\|^2,
\end{equation}
where $\bfD(\bfx) = \begin{bmatrix} \frac1{Z(\bfx)}\bfA(\bfx) &\bfB(\bfx) \end{bmatrix}$ is the $2\times 6$ feature sensitivity matrix \cite{Corke17book, Bryner19icra}.
The minimal problem in this context requires six normal flow measurements to determine the 6-DoF motion parameters.

\subsubsection{Motion and Structure Estimation in Planar Scenes.}
\label{subsubsec:motion_estimation_in_planar_scenes}
For a moving camera that observes a planar scene, the motion and structure parameters are compatibly encoded in a homography matrix.
In our case, we attempt to recover the differential homography $\bfh_d$ (the vector form of $\bfH_{d}$ in \eqref{eq:differ_homo}) because the instantaneous linear and angular velocities are considered.
By expanding \eqref{eq:differ_homo} and rearranging equations, we have
\begin{equation}
    \label{eq:e_diff_homo}
    \bfu(\bfx)= \bfC(\bfx) \bfh_d,
\end{equation}
where 
$ \bfC(\bfx)$ is a $2\times 9$ matrix that solely depends on the pixel location $\bfx$.
Subsequently, substituting \eqref{eq:normalflow:dotprod} into \eqref{eq:e_diff_homo}, gives:
\begin{equation}
\bfn(\bfx)^{\top} \bfC(\bfx) \bfh_d = \|\bfn(\bfx)\|^2.
\end{equation}
As discussed in \cite{Differential_Homo_ECCV20}, retrieving the motion and structure parameters from the decomposition of $\bfh_d$ is not trivial and beyond the scope of this paper.
Therefore, we only focus on the numerical solution of $\bfh_d$.

\subsection{Our Solvers}
\label{subsec:our solvers}

According to the motion-and-structure models derived in \cref{subsec:models of motion ans structure}, we have established certain constraints between the model parameters and the normal flow observations.
This allows us to simply develop linear solvers for each specific problem.
Besides, we further propose a continuous-time nonlinear solver that can effectively handle the presence of sudden variations in motion and does not rely on the constant-motion assumption.
To convey our solvers concisely, we generalize the motion-and-structure models into a unified form, as
\begin{equation}
\label{eq:general_motion_flow_model}
\bfn(\bfx)^{\top} \bfO(\bfx) \bparams = \|\bfn(\bfx)\|^2,
\end{equation}
where $\bfO(\bfx)$ is the system matrix\footnote{$\bfO(\bfx)$ degenerates to 1 in the problem of optical flow estimation.} and $\bparams$ are the motion/structure parameters.

\subsubsection{Linear Solver.}
\label{subsubsec:batch_based_instantaneous_solver}
Given as input a set (\eg, $\mathrm{K}$) of sparse normal-flow estimates obtained at the same time\footnote{An approximation can be made based on the assumption of constant camera motion.
In this way, the normal flow estimates can be collected within a short time interval.}, our linear solver works out the following (overdetermined) linear system of equations:
\begin{equation}
\label{eq:linear_least_squares}
\begin{pmatrix}
\bfn(\bfx_{1})^{\top} \bfO(\bfx_{1})\\
\vdots\\
\bfn(\bfx_{\mathrm{K}})^{\top} \bfO(\bfx_{\mathrm{K}})
\end{pmatrix}
\bparams= 
\begin{pmatrix}
\|\bfn(\bfx_{1})\|^2\\
\vdots\\
\|\bfn(\bfx_{\mathrm{K}})\|^2
\end{pmatrix}.
\end{equation}
For robust estimation, RANSAC \cite{Fischler81cacm} is adopted to deal with noise and outliers.

\subsubsection{Continuous-Time Nonlinear Solver.}
\label{subsubsec:asynchronous_continuous_time_solver}
To handle the case of a sudden variation in speed that may violate the constant-motion assumption, we further propose a continuous-time nonlinear solver.
In particular, we employ cubic B-splines to represent the continuous-time model parameters as
\begin{equation}
\label{eq:b-spline formula of velocity}
    \os{}{}{\bparams}{}{}(u(t)) = \sum_{i=0}^{3}B_{i,3}(u(t)) \os{}{}{\bfc}{}{i},
\end{equation}
where $B_{i,k}$ denotes the basis function, $i$ the index of control points, $k$ (=3) the order of the spline and $\os{}{}{\bfc}{}{i} \in \Real^N$ the corresponding control point defined in the $N$-dimensional space of model parameters.
Examples of cubic B-splines applied to event camera motion estimation are found in \cite{Mueggler18tro,Guo24tro}.
The function $u(\cdot)$ is a normalization operator, which transfers time $t$ to the spline's parameter domain by means of basis translation~\cite{de1972calculating}.
In this way, the model parameters at any time instant $t$ can be obtained via an interpolation using the known control points.

The formulation of the continuous-time model fitting problem goes as follows.
The unknown variable consists of all the involved control points, $\boldsymbol{\cX} \doteq [\bfc_0, \bfc_1, \cdots \bfc_n]$,
where $n$ denotes the total number of control points.
By introducing the continuous-time representation of model parameters to \eqref{eq:general_motion_flow_model}, we finally create the following objective function:
\begin{equation}
\boldsymbol{\cX}^{\ast}
= \arg\min_{\boldsymbol{\cX}} \sum_{e_k \in \mathcal{E}}^{} 
\left\|
\bfn(\bfx_{k})^{\top} \bfO(\bfx_{k}) \os{}{}{\bparams}{}{}(u(t_k)) - \|\bfn(\bfx_{k})\|^2 \right\|^2,
\end{equation}
where $\mathcal{E}$ denotes the event set in which the norm flow observation is available for every involved event $e_k$.
Note that this continuous-time formulation aligns better with the asynchronous nature of event-based normal flow, enabling a state estimation over a larger temporal window that may lead to a better signal-to-noise ratio.
The linear solver can be used to initialize the continuous-time nonlinear solver.
An illustrative explanation of our solvers is given in \cref{fig:our_slovers}.

\section{Experiments}
\label{sec:evaluation}

To justify the effectiveness of the proposed algorithms, experiments are conducted on both simulation and real data. 
First, we introduce the datasets used and the applied metrics for evaluation (\cref{subsec:datasets and evaluation metrics}). 
Then we report the results of our linear solver and continuous-time nonlinear solver (\cref{subsec:evaluation of linear solver} and \cref{subsec:evaluation of continuous-time solver}).
Finally, a discussion on the computational complexity and the limitation of our method is provided in \cref{subsec:discussion}.

\subsection{Experimental Setup}
\label{subsec:datasets and evaluation metrics}
Our evaluation utilizes four datasets, including the indoor event-camera dataset (ECD) captured with a DAVIS240C camera and motion-capture system for groundtruth, and a high-resolution dataset with two rotational motion sequences using an iniVation DAVIS346 camera and Xsens IMU. The VECtor dataset, captured with a Prophesee Gen 3 camera and an accompanying RGB-D depth camera, is used for 6-DoF motion tracking. 
Additionally, a synthetic dataset generated in Blender is employed for motion-and-structure pattern analysis. 
Evaluation metrics for kinematics estimation include average error (AE) and root mean squared error (RMSE), while differential homography estimation is assessed using the Frobenius norm. 
Best results are in bold, and second-best are underlined. 
The supplementary material provides detailed descriptions of datasets and metrics, as well as implementation details of our methods.

\begin{figure}[t]
    \centering
    \begin{subfigure}{0.34\linewidth}
    \frame{\includegraphics[width=\textwidth]{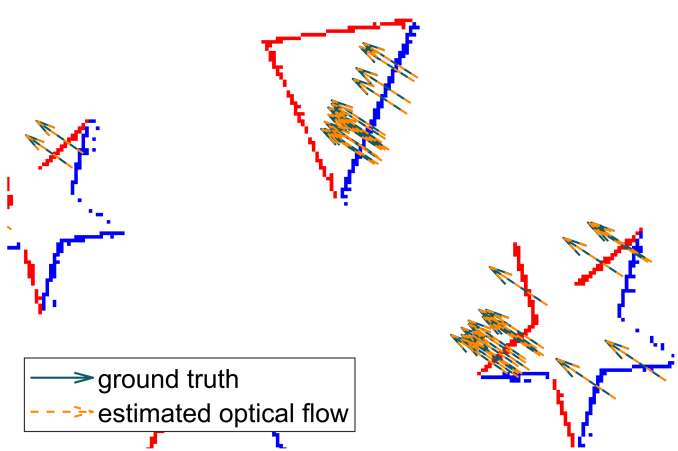}}
    \caption{Estimated optical flow over a map of event accumulation.}
    \label{subfig:oflow}
    \end{subfigure}
    \;
    \begin{subfigure}{0.63\linewidth}
    \includegraphics[width=\textwidth]{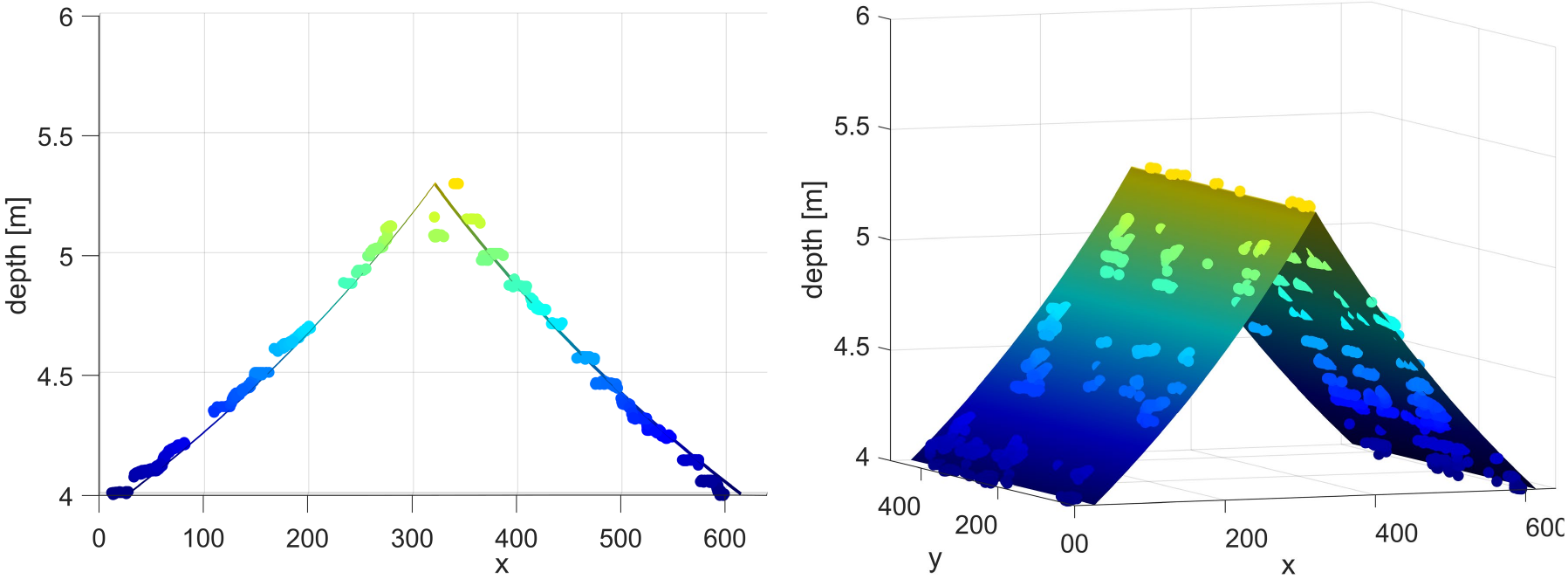}
    \caption{Reconstruction result displayed over groundtruth structures (a v-shape roof) and viewed from different perspectives.}
    \label{subfig:depth}
    \end{subfigure}
    \caption{Qualitative result of optical flow and depth on sequence \emph{three\_wall\_translation} of our synthetic data.
    Note that our method returns sparse results.}
    \label{fig:oflow and depth estimation}
\end{figure}

\subsection{Evaluation of the Linear Solver}
\label{subsec:evaluation of linear solver}
In the following evaluation, we qualitatively and quantitatively assess the performance of our linear method (Ours) in each specific task, and compare against different state-of-the-art (SOTA) methods (CMax\cite{Gallego18cvpr} and Pro-STR\cite{huang2023progressive}) and their customized version initialized using our linear solver.

\subsubsection{Optical Flow and Depth Estimation.}
\label{subsec:experiment_optical_flow}

Due to a lack of benchmark for extensive evaluation, we qualitatively evaluate our method on the tasks of optical flow and depth estimation.
As illustrated in \cref{subfig:oflow}, our linear solver demonstrates the capability to accurately reconstruct full flow using normal flow and known motion parameters.
An example of the depth estimation result is given in \cref{subfig:depth}, showing that the recovered structure aligns well with the groundtruth.

\subsubsection{Angular Velocity Estimation.}
\label{subsec:experiment_rotation}

As shown in \cref{tab:2}, we see that our linear solver outperforms CMax and Pro-STR.
Besides, we find the combination of our linear solver and CMax leads to the best result, indicating that the proposed linear solver is a complementary method to existing nonlinear methods.

\begin{table}[htbp]
\caption{Evaluation of our linear solver on the task of angular velocity estimation.}
\label{tab:2}
\centering
\adjustbox{max width=\linewidth}{
\newcommand{\tabincell}[2]{\begin{tabular}{@{}#1@{}}#2\end{tabular}}
\setlength{\tabcolsep}{0.006\linewidth}
\begin{tabular}{l|cc|cc|cc}
\Xhline{1pt}
\multirow{3}{*}{\textbf{Method}}
&\multicolumn{2}{c|}{\emph{shapes\_rotation}\cite{Mueggler17ijrr}}
&\multicolumn{2}{c|}{\emph{ground\_rotation} (our data)}
&\multicolumn{2}{c}{\emph{patterns\_rotation} (our data)}\\
\cline{2-7}
&\multirow{2}{*}{\tabincell{c}{$e_{\omega}$\\(deg/s)}}
&\multirow{2}{*}{\tabincell{c}{$\text{RMSE}_{\omega}$\\(deg/s)}}
&\multirow{2}{*}{\tabincell{c}{$e_{\omega}$\\(deg/s)}}
&\multirow{2}{*}{\tabincell{c}{$\text{RMSE}_{\omega}$\\(deg/s)}}
&\multirow{2}{*}{\tabincell{c}{$e_{\omega}$\\(deg/s)}}
&\multirow{2}{*}{\tabincell{c}{$\text{RMSE}_{\omega}$\\(deg/s)}}\\
&   &   &   &   &   &  \\
\Xhline{0.7pt}
CMax\cite{Gallego18cvpr} 
&15.60  &26.38  
&9.48   &15.58  
&5.95   &9.07\\
Pro-STR\cite{huang2023progressive} 
&90.61  &145.81 
&31.07  &44.74  
&10.47  &11.88\\
Ours
&\underline{15.36}  &\underline{23.98}  
&\underline{3.30}   &\underline{8.01} 
&4.70               &6.08\\
Ours+CMax\cite{Gallego18cvpr} 
&\textbf{10.12}     &\textbf{14.29} 
&\textbf{2.85}      &\textbf{3.74}  
&\textbf{0.35}      &\textbf{0.73}\\
Ours+Pro-STR\cite{huang2023progressive}   
&19.23              &40.44  
&9.13               &17.07  
&\underline{4.31}   &\underline{5.23}\\
\Xhline{1pt}
\multirow{3}{*}{\textbf{Method}}
&\multicolumn{2}{c|}{\emph{dynamic\_rotation}\cite{Mueggler17ijrr}}
&\multicolumn{2}{c|}{\emph{boxes\_rotation} (our data)}
&\multicolumn{2}{c}{\emph{cubes\_rotation} (our data)}\\
\cline{2-7}
&\multirow{2}{*}{\tabincell{c}{$e_{\omega}$\\(deg/s)}}
&\multirow{2}{*}{\tabincell{c}{$\text{RMSE}_{\omega}$\\(deg/s)}}
&\multirow{2}{*}{\tabincell{c}{$e_{\omega}$\\(deg/s)}}
&\multirow{2}{*}{\tabincell{c}{$\text{RMSE}_{\omega}$\\(deg/s)}}
&\multirow{2}{*}{\tabincell{c}{$e_{\omega}$\\(deg/s)}}
&\multirow{2}{*}{\tabincell{c}{$\text{RMSE}_{\omega}$\\(deg/s)}}\\
&   &   &   &   &   &  \\
\Xhline{0.7pt}
CMax\cite{Gallego18cvpr}
&16.14  &26.48  
&8.18   &13.56 
&6.77    &10.62\\
Pro-STR\cite{huang2023progressive}
&54.64  &85.63 
&22.72  &32.05  
&11.07  &12.79\\
Ours
&\underline{9.41}   &\underline{18.06}  
&\textbf{2.60}      &\underline{4.78}   
&2.31               &3.02\\
Ours+CMax\cite{Gallego18cvpr}
&\textbf{7.71}      &\textbf{10.85} 
&\underline{2.64}   &\textbf{3.80}      
&\textbf{0.49}      &\textbf{0.77}\\
Ours+Pro-STR\cite{huang2023progressive}
&18.87              &32.65  
&3.49               &5.87   
&\underline{2.04}   &\underline{2.76}\\
\Xhline{1pt}
\end{tabular}
}
\end{table}

\begin{table}[htbp]
\caption{Evaluation of our linear solver for differential homography estimation.}
\label{tab:homo}
\centering
\adjustbox{max width=0.75\linewidth}{
\setlength{\tabcolsep}{0.01\linewidth}
\begin{tabular}{l|c|c|c}
\Xhline{1pt}
\multirow{2}{*}{\textbf{Method}}
&\emph{board\_slow}\cite{Gao22ral}     
&\emph{patterns\_rotation}    
&\emph{patterns\_6dof}\\
\cline{2-4}
&F-Norm
&F-Norm
&F-Norm\\
\Xhline{0.7pt}
CMax\cite{Gallego18cvpr}
& 1.16    
& 1.33    
& 1.27    \\
Pro-STR\cite{huang2023progressive}
& 2.07    
& 1.22    
& 1.37   \\
Ours
& 0.24 
& 0.20 
& 0.43 \\
Ours+CMax\cite{Gallego18cvpr}
& \underline{0.23}  
&\textbf{0.15}   
&\textbf{0.41}   \\
Ours+Pro-STR\cite{huang2023progressive}
& \textbf{0.19}   
& \underline{0.17}  
& \underline{0.42}  \\
\Xhline{1pt}
\end{tabular}
}
\end{table}

\begin{table}[htbp]
\caption{Evaluation of our linear solver on the task of 6-DoF motion tracking.}
\label{tab:4}
\centering
\adjustbox{max width=\linewidth}{
\newcommand{\tabincell}[2]{\begin{tabular}{@{}#1@{}}#2\end{tabular}}
\setlength{\tabcolsep}{0.015\linewidth}
\begin{tabular}{l|cccc|cccc}
\Xhline{1pt}
\multirow{3}{*}{\textbf{Method}}
&\multicolumn{4}{c|}{\emph{corner\_slow}\cite{Gao22ral}}
&\multicolumn{4}{c}{\emph{patterns\_6dof} (our data)}\\
\cline{2-9}
&\multirow{2}{*}{\tabincell{c}{$e_{\omega}$\\(deg/s)}}
&\multirow{2}{*}{\tabincell{c}{$\mathrm{RMSE}_{\omega}$\\(deg/s)}}
&\multirow{2}{*}{\tabincell{c}{$e_{\upsilon}$\\(m/s)}}
&\multirow{2}{*}{\tabincell{c}{$\mathrm{RMSE}_{\upsilon}$\\(m/s)}}
&\multirow{2}{*}{\tabincell{c}{$e_{\omega}$\\(deg/s)}}
&\multirow{2}{*}{\tabincell{c}{$\mathrm{RMSE}_{\omega}$\\(deg/s)}}
&\multirow{2}{*}{\tabincell{c}{$e_{\upsilon}$\\(m/s)}}
&\multirow{2}{*}{\tabincell{c}{$\mathrm{RMSE}_{\upsilon}$\\(m/s)}}
\\
&   &   &   &   &   &   &   &   \\

\Xhline{0.7pt}

CMax\cite{Gallego18cvpr}
& 12.39    & 14.39 & 0.34   & 0.41 
 & 12.94  & 15.68 & 1.02 & 1.23   \\

Pro-STR\cite{huang2023progressive}
&5.19    & 7.84  & 0.13   & 0.22 
 & 11.21  & 13.49 & 1.33 & 1.49   \\
Ours
& \underline{1.31}   & \underline{1.79}   & \textbf{0.01}    & \textbf{0.02} 
&\underline{0.56}   &\underline{0.89}   &\textbf{0.12}  &\underline{0.16}
 \\
Ours+CMax\cite{Gallego18cvpr}
&2.19   &2.82   &0.11   &0.17   
&\textbf{0.55}  &\textbf{0.87}  &0.18   &0.30   
   \\

Ours+Pro-STR\cite{huang2023progressive} 
& \textbf{1.06}   & \textbf{1.34}  &\underline{0.04}  &\underline{0.05} 
&1.07   &1.56   &\textbf{0.12}  &\textbf{0.15}  
      \\
\Xhline{1pt}
\multirow{3}{*}{\textbf{Method}}
&\multicolumn{4}{c|}{\emph{mountain\_normal}\cite{Gao22ral}}
&\multicolumn{4}{c}{\emph{cubes\_6dof} (our data)}\\
\cline{2-9}
&\multirow{2}{*}{\tabincell{c}{$e_{\omega}$\\(deg/s)}}
&\multirow{2}{*}{\tabincell{c}{$\mathrm{RMSE}_{\omega}$\\(deg/s)}}
&\multirow{2}{*}{\tabincell{c}{$e_{\upsilon}$\\(m/s)}}
&\multirow{2}{*}{\tabincell{c}{$\mathrm{RMSE}_{\upsilon}$\\(m/s)}}
&\multirow{2}{*}{\tabincell{c}{$e_{\omega}$\\(deg/s)}}
&\multirow{2}{*}{\tabincell{c}{$\mathrm{RMSE}_{\omega}$\\(deg/s)}}
&\multirow{2}{*}{\tabincell{c}{$e_{\upsilon}$\\(m/s)}}
&\multirow{2}{*}{\tabincell{c}{$\mathrm{RMSE}_{\upsilon}$\\(m/s)}}
\\
&   &   &   &   &   &   &   &   \\
\Xhline{0.7pt}
CMax\cite{Gallego18cvpr}
&11.01  &14.17  &0.44   &0.52
& 10.74  & 14.33 & 0.78 & 1.05 \\
Pro-STR\cite{huang2023progressive}
&5.23   &10.53  &0.14   &0.25
& 9.75  & 11.40 & 1.29 & 1.44\\
Ours
&\underline{1.28}   &\underline{1.83}   &\underline{0.07}   &\textbf{0.09} 
&\textbf{0.79}      &\textbf{1.02}      &\textbf{0.11}  &\textbf{0.14}      \\
Ours+CMax\cite{Gallego18cvpr}
&3.84   &6.30   &0.21   &0.30
&1.14   &1.72   &0.12   &0.14  \\
Ours+Pro-STR\cite{huang2023progressive}
&\textbf{1.18}      &\textbf{1.58}      &\textbf{0.06}      &\textbf{0.09}
&\underline{1.05}   &\underline{1.33}   &\textbf{0.11}   &\textbf{0.14}  \\
\Xhline{1pt}
\end{tabular}
}
\end{table}

\begin{figure}
  \centering
  \begin{subfigure}{0.92\linewidth}
    \includegraphics[width=\textwidth]{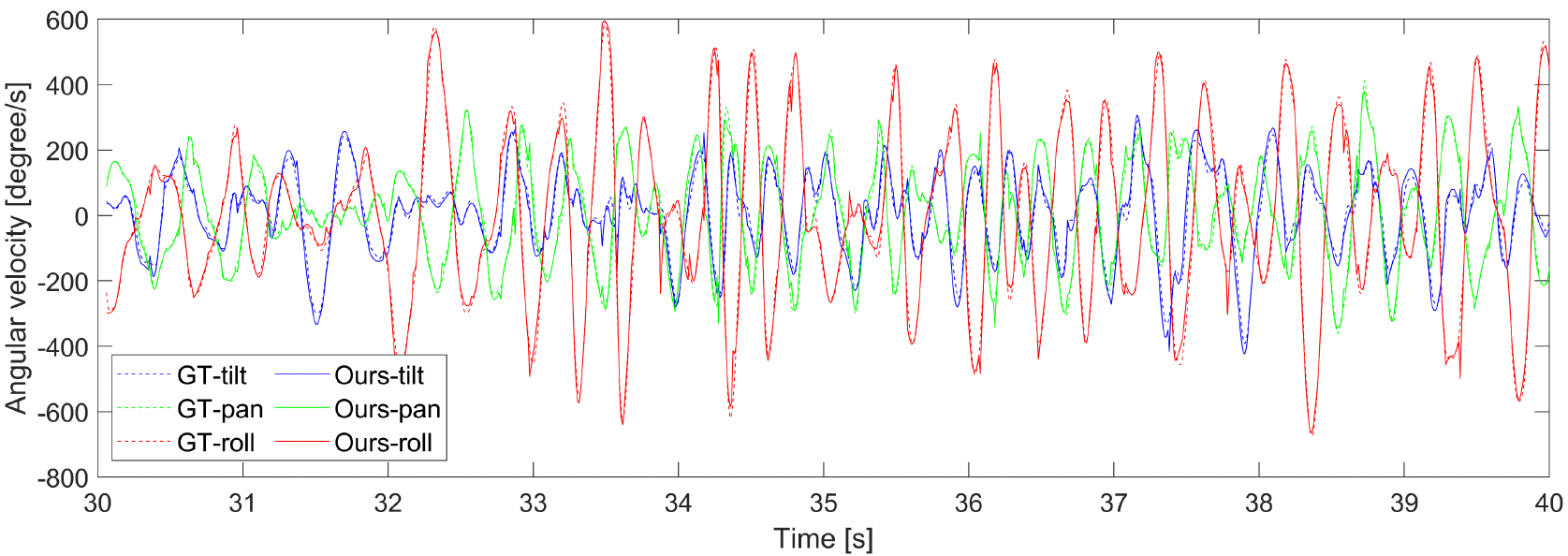}
    \caption{Results of the continuous-time angular velocity estimator.}
    \label{fig:bspline}
  \end{subfigure}
\begin{subfigure}{0.92\linewidth}
\includegraphics[width=\textwidth]{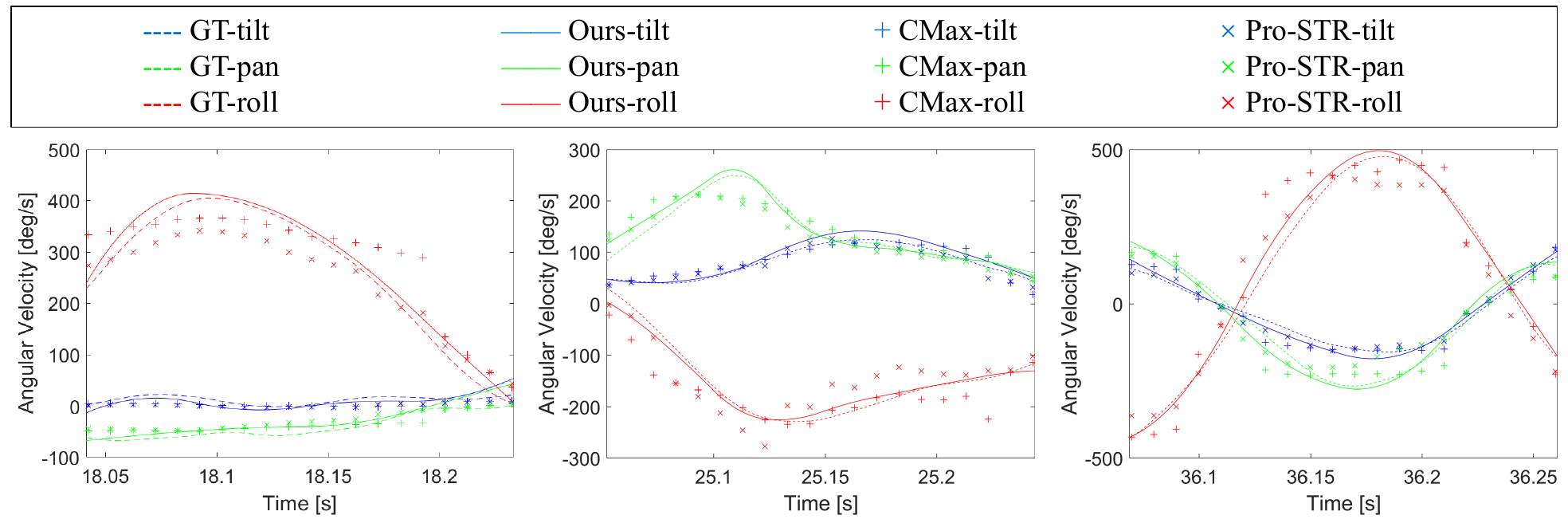}
\caption{Comparison against two state-of-the-art nonlinear solutions using data segments that involve sudden variations of angular velocity.}
\label{fig:NACTS_1}
\end{subfigure}
\caption{Angular velocity estimation using the continuous-time nonlinear solver (Ours).}
  \label{fig:shapes_rotation_results}
\end{figure}

\subsubsection{Differential Homography Estimation.}
\label{subsec:experiment_homo}

The result is consistent with that in the above task.
Table~\ref{tab:homo} reveals that our
linear method outperforms CMax and Pro-STR in terms of numerical accuracy.
The combination of our linear solver and state-of-the-art nonlinear solver leads to the best result.

\subsubsection{6-DoF Motion Tracking.}
\label{subsec:experiment_tracking}

Given known depth, we recover the linear and angular camera velocities on the task of 6-DoF motion estimation.
As shown in \cref{tab:4}, our linear solver reports more accurate results than both CMax and Pro-STR.
In addition, we surprisingly find that the linear solver can sometimes outperform the combination of itself with the two nonlinear solvers.
This is, according to our investigation, attributed to the violation of constant-motion assumption in some sequences, which is held for the two nonlinear solvers.

\subsection{Evaluation of the Continuous-Time Solver}
\label{subsec:evaluation of continuous-time solver}

To justify the claimed advantage of the proposed continuous-time nonlinear solver, we use the exemplary task of angular velocity estimation. 
The sequence used (\emph{shapes\_rotation} \cite{Mueggler17ijrr}) captures a purely rotating event camera, and some parts of the data involves aggressive motion.
As illustrated in \cref{fig:bspline}, our nonlinear solver gives accurate estimation results with a high stability.
Moreover, we select three segments that witness a sudden variation of angular speed within a short time interval and compare our continuous-time nonlinear solver against CMax and Pro-STR.
Note that all of them are initialized by our linear method for a fair comparison.
As \cref{fig:NACTS_1} shows, our results are more consistent with the groundtruth angular velocity than the other two methods, which are less accurate due to the dependence on the constant-motion assumption.

\subsection{Discussion}
\label{subsec:discussion}

\subsubsection{Computational Complexity.}
The computational complexity of the linear solver depends on the specific task of model fitting.
For optical flow and depth estimation, it is linearly proportional to the size of the input event flow, and thus, the complexity is $\mathcal{O}(N)$.
For the other tasks, the time cost predominantly hinges on the total number of RANSAC iterations and the computations performed per iteration.
In each iteration, we typically solve a linear overdetermined system via QR decomposition at a complexity of $\mathcal{O}(c N^2)$, where $c$ is the cardinality of a minimal problem (\eg, $c=3$ for the problem of angular velocity estimation).

\subsubsection{Runtime.}
{All experiments are conducted on a standard computer with Intel i7-13700F CPU, 16G memory and implemented with MATLAB (version R2022a).
Using the 6-DoF tracking task as an example, we measure the runtime of the proposed linear solver and continuous-time (CT) solver, as reported in \cref{tab:time_cost}.}
\begin{table}[t]
\caption{Runtime of 6-DoF tracking on sequence \emph{mountain\_normal} (640$\times$480 px).}
\label{tab:time_cost}
\centering
\adjustbox{max width=0.80\linewidth}{
\setlength{\tabcolsep}{6pt}
\begin{tabular}{l|l}
\Xhline{1pt}
\textbf{Component}      &\textbf{Runtime}\\
\Xhline{0.7pt}
Normal flow extraction  &\multicolumn{1}{c}{ 20 ms (for extracting 2000 normal flow vectors)}\\
Linear solver           &25.97 ms   \\
CT solver               &28.29 ms   \\
\Xhline{1pt}
\end{tabular}
}
\end{table}

\subsubsection{Limitations.}
To make it clear, we emphasize that our method, like other compared methods, does not resolve the motion and structure parameters simultaneously.
It should also be noted that the success of our method relies heavily on the quality of the input normal flow.
Particularly, our normal flow extractor (including some other open source solutions for computing normal flow from event data) may fail in the presence of densely repetitive textures, leading to a failure of the modeling fitting scheme.
To assess the sensitivity of our method to the quality of event-based normal flow, we propose an analysis on numerical stability using simulation data in the supplementary material.

\section{Conclusions}
\label{sec:conclusions}

This work presents a solution to the general problem of geometric model fitting with event data.
Considering the differential working principle of event cameras, our method establishes specific constraints between instantaneous motion-and-structure parameters and event-based normal flow.
Experiments show that the proposed linear solver leads to closed-form and deterministic solutions, which can be used to initialize existing state-of-the-art nonlinear methods.
The proposed continuous-time nonlinear solver, furthermore, demonstrates the capability to handle sudden speed variations, obviating the dependence on the constant-motion assumption.
Finally, we hope this work can inspire future research on event-based geometric model fitting, a paramount topic to many downstream techniques, such as event-based 3-D vision, SLAM, and scene segmentation.

\section*{Acknowledgments}
We thank Dr. Yi Yu for proof reading.
This work was supported by the National Key Research and Development Project of China under Grant 2023YFB4706600.
Funded by the Deutsche Forschungsgemeinschaft (DFG, German Research Foundation) under Germany’s Excellence Strategy – EXC 2002/1 ``Science of Intelligence'' – project number 390523135.

\appendix
\section*{SUPPLEMENTARY MATERIAL}
\section*{Overview}
In this supplementary material we provide the following:
\begin{itemize}
    \item A detailed review of preliminaries involved, including motion field, differential geometry and event-based normal flow (Sec.~\ref{sec:preliminaries_supp}).
    \item Additional details about differential homography (Sec.~\ref{sec:method_supp}).
    \item Details of dataset and evaluation metrics utilized (Sec.~\ref{sec:dataset_supp}).
    \item Implementation details of the proposed algorithms (Sec.~\ref{sec:implementation_supp}).
    \item More evaluations on the proposed methods, including an analysis of numerical stability, and the performance of the continuous-time nonlinear solver in textured scenes. (Sec.~\ref{sec:extensive_experiments_supp}).
\end{itemize}

\section{Supplement to Preliminaries}
\label{sec:preliminaries_supp}
We supplement Sec.~3 of the main paper by providing a detailed review on three important concepts, including \emph{motion field} (Sec.~\ref{subsec:motion_flow_and_normal_flow}), \emph{differential geometry} across multiple views (Sec.~\ref{subsec:multi_view_differential_geometry}), and \emph{event-based normal flow} (Sec.~\ref{subsec:event_based_normal_flow}).

\subsection{Motion Field}
\label{subsec:motion_flow_and_normal_flow}
Consider a visual observer moving in a static environment, and let its instantaneous angular velocity and linear velocity be denoted in the observer's body frame as $\os{\ttB}{}\angvel{}{}=(\omega_{x},\omega_{y},\omega_{z})^{\top}$ and $\os{\ttB}{}\linvel{}{}=(\nu_{x},\nu_{y},\nu_{z})^{\top}$, respectively.
The scene flow of a 3D point $\os{\ttB}{}\bfP{}{} = (X,Y,Z)^{\top}$ is
\begin{align}
\label{eq:supp:scene_flow}
\dot{\os{\ttB}{}\bfP{}{}} = -\os{\ttB}{}\angvel{}{} \times \os{\ttB}{}\bfP{}{} - \os{\ttB}{}\linvel{}{}.
\end{align}
Let $\bfx = (x,y)^{\top} = (X/Z, Y/Z)^{\top}$ be the image of $\bfP$ represented in calibrated camera coordinates.
Projecting the scene flow vector onto the image plane gives us the relationship between the optical flow $\bfu$ (in calibrated coordinates) and the dynamics of the observer as
\begin{align}
&\bfu = \frac{1}{Z}\mathbf{A}(\bfx)\,\os{\ttB}{}\linvel{}{} + \mathbf{B}(\bfx)\,\os{\ttB}{}\angvel{}{},
\label{eq:supp:motion_flow_equation}\\ 
& \mathbf{A}(\bfx) = \begin{bmatrix}
-1 & 0 & x\\ 
0 & -1 & y
\end{bmatrix},\\
& \mathbf{B}(\bfx) = \begin{bmatrix}
x y & -(1+{x}^2) & y\\ 
(1+{y}^2) & -x y & -x
\end{bmatrix}, 
\end{align}
where \eqref{eq:supp:motion_flow_equation} is acknowledged as the motion field equation \cite{longuet1980interpretation}.
The depth $Z$ is different for every pixel $\bfx$, hence we often write $Z=Z(\bfx)$. 

\subsection{Multi-View Differential Geometry}
\label{subsec:multi_view_differential_geometry}
The theory of differential geometry across multiple views can be derived by taking the temporal derivatives of the standard multi-view geometry \cite{Ma04book}.
It examines the relationship between the optical flow and the instantaneous kinematics across multiple views.
We focus on two specific two-view geometric properties: 
1) the differential homography~\cite{Differential_Homo_ECCV20}, and 
2) the differential epipolar geometry~\cite{Differertial_RS_ICCV17}. 

\subsubsection{Differential Homography.}
The standard homography is a projective transformation that describes the image motion between two views of a camera observing a planar scene.
Such a mapping function, induced by the planar scene, can be expressed as
\begin{equation}
    \label{eq:supp:homo}
    \hat{\bfx}^{\prime} = \bfH\hat{\bfx},
\end{equation}
where $\hat{\bfx}$ and $\hat{\bfx}^{\prime}$ denote the corresponding pixels in terms of homogeneous calibrated image coordinates, and $\bfH \in \Real^{3\times3}$ is the homography matrix. 
The induced homography is a function of the relative pose $\{\Rot, \bft\}$ and the structure parameters $\{\bfN, d\}$ (i.e., parameters of the plane in the scene), and can be expressed as $\bfH \doteq \Rot - \frac{1}{d}\bft\bfN^{\top}$.
The differential version replaces the potentially large displacement of the transformation $\bfx \mapsto \bfx'$ by its first-order continuous-time approximation $\bfx \mapsto \bfx + \Delta t \hat{\bfu}$, with point velocity given by:
\begin{equation}
    \label{eq:supp:differ_homo}
    \hat{\bfu}(\bfx) = \left(\bfI - \hat{\bfx}\bfe_3^{\top}\right)\bfH_d\hat{\bfx}.
\end{equation}
Here, $\bfH_{d} \doteq -\left([\angvel]_{\times} + \frac1{d}\linvel\bfN^{\top}\right)$ denotes the differential homography matrix ($[\cdot]_\times$ denotes the cross-product, skew-symmetric matrix), $\bfe_3 = (0, 0, 1)^{\top}$, $\hat{\bfx}= (x, y, 1)^{\top}$, and $\hat{\bfu}= (u_x, u_y, 0)^{\top}$ the optical flow (in homogeneous coordinates).

\subsubsection{Differential Epipolar Geometry.}
As a relatively relaxed constraint compared to the homography, the standard epipolar geometry only defines a potential direction for searching the correspondence pair $\hat{\bfx} \leftrightarrow \hat{\bfx}^{\prime}$.
Such a constraint evaluates the distance between a potential match to the determined epipolar line, and can be expressed as
\begin{equation}
    \label{eq:supp:epipolar}
    \hat{\bfx}^{\prime\top}\bfE\hat{\bfx} = 0.
\end{equation}
The differential epipolar geometry can be derived similarly, yielding
\begin{equation}
    \label{eq:supp:diff_epi}
    \hat{\bfu}(\bfx)^{\top}[\boldsymbol{\nu}]_\times \hat{\bfx} - \hat{\bfx}^{\top} \bfs \hat{\bfx} = 0,
\end{equation}
where $\bfs = \frac{1}{2}([\boldsymbol{\nu}]_\times[\angvel]_\times+[\angvel]_\times [\boldsymbol{\nu}]_\times
)$. 
Note that \eqref{eq:supp:diff_epi} can be derived directly from \eqref{eq:supp:motion_flow_equation} by eliminating the depth information.

\subsection{Event-Based Normal Flow}
\label{subsec:event_based_normal_flow}

\subsubsection{Normal Flow.}
\label{subsubsec:normal_flow}

While the motion field is the true projection of the 3D scene velocities on the image plane, 
its determination from visual quantities measured at the image plane is called ``optical flow''. 
The optical flow is the apparent motion of patterns on the image plane. 
As is well known, the optical flow can only be a good approximation for the true motion (i.e., the motion field) at image points that are unambiguous to track, i.e., that are surrounded by edge patterns that accurately determine their location in both image plane directions ($x$ and $y$).
A common assumption that is leveraged to compute optical flow is ``brightness constancy'', that is, the approximation that image brightness remains constant as the image point $\bfx$ moves. This is an approximation for points corresponding to matt and Lambertian objects; it is not a sensible assumptions for points on shiny and specular objects.

Letting $I(\bfx,t)$ be the brightness function on the image plane as a function of time, 
the brightness constancy assumption can be compactly written as:
\begin{equation}
I(\bfx(t),t)=\text{const}.
\end{equation}
A first-order Taylor expansion gives
\begin{equation}
I(\bfx+\Delta\bfx, t+\Delta t) \approx I(\bfx,t) + (\nabla I(\bfx,t))^{\top} \dot{\bfx}\Delta t + \partial_tI(\bfx,t) \Delta t,
\end{equation}
where $\dot{\bfx} \equiv \dot{\bfx}(t) = (dx/dt, dy/dt)^\top$ is the velocity of image point $\bfx(t)$, 
$\nabla = (\partial_x, \partial_y)^\top$ are the spatial derivatives and $\partial t$ is the temporal derivative.
Moving terms around,
\begin{equation}
I(\bfx+\Delta\bfx, t+\Delta t) - I(\bfx,t) \approx (\nabla I(\bfx,t))^{\top} \dot{\bfx}\Delta t + \partial_tI(\bfx,t) \Delta t.
\end{equation}
Brightness constancy states that for the true motion curves $\bfx(t)$, the intensity remains the same, 
$I(\bfx+\Delta\bfx, t+\Delta t) \approx I(\bfx,t)$, and therefore (since $\Delta t >0$)
\begin{equation}
\label{eq:supp:brightness_consistency_equation}
(\nabla I(\bfx,t))^{\top} \dot{\bfx} + \partial_tI(\bfx,t) \approx 0.
\end{equation}
Assuming that the brightness $I$ is known (i.e., measured by a camera), 
\eqref{eq:supp:brightness_consistency_equation} is one equation in two unknowns ($\dot{\bfx} \in \Real^2$).
More equations are needed to be able to determine the velocity $\dot{\bfx}$ at point $(\bfx,t)$ in space-time.
Without additional equations or information, all that can be determined is the velocity component that is parallel to $\nabla I(\bfx,t)$ (i.e., perpendicular to the edge) due to the dot product operation.
That is, decomposing $\dot{\bfx} = \dot{\bfx}_{\parallel} + \dot{\bfx}_{\perp}$
into its parallel and perpendicular components to the local edge $\nabla I$, respectively,
we have $\nabla I(\bfx,t)^{\top} \dot{\bfx}_{\parallel} = 0$.
Substituting in \eqref{eq:supp:brightness_consistency_equation} gives:
\begin{eqnarray}
0 &\approx& 
(\nabla I(\bfx,t))^{\top} \left( \dot{\bfx}_{\parallel} + \dot{\bfx}_{\perp} \right) + \partial_tI(\bfx,t)\\
&=& \underbrace{(\nabla I(\bfx,t))^{\top} \dot{\bfx}_{\parallel}}_{0} 
+ (\nabla I(\bfx,t))^{\top} \dot{\bfx}_{\perp} + \partial_tI(\bfx,t) \\
&=& (\nabla I(\bfx,t))^{\top} \dot{\bfx}_{\perp} + \partial_tI(\bfx,t) \label{eq:supp:BrightnessConstDifferential}
\end{eqnarray}

From here, we can work out $\bfn \equiv \dot{\bfx}_{\perp} \propto \nabla I$.
We know that 
\begin{equation} 
\bfn \equiv \dot{\bfx}_{\perp} = \|\dot{\bfx}_{\perp}\| \frac{\nabla I}{\|\nabla I\|}.\label{eq:supp:normalflowdef}
\end{equation}
Substituting in \eqref{eq:supp:BrightnessConstDifferential} (and omitting the evaluation point $(\bfx,t)$, for clarity),
\begin{equation}
(\nabla I)^{\top} \|\dot{\bfx}_{\perp}\| \frac{\nabla I}{\|\nabla I\|} = -\partial_t I,
\end{equation}
which gives
\begin{equation}
\|\dot{\bfx}_{\perp}\| = - \partial_t I \frac{ \|\nabla I\| }{(\nabla I)^{\top}\nabla I} 
= - \partial_t I \frac1{ \|\nabla I\| }.
\end{equation}
Substituting in \eqref{eq:supp:normalflowdef}, it finally gives
\begin{equation}
\label{eq:supp:normal_flow_equation_1}
    \bfn(\bfx,t) = -\frac{\partial_t I(\bfx,t)}{\Vert \nabla I(\bfx,t) \Vert^2} \nabla I(\bfx,t).
\end{equation}
This component of the optical flow $\dot{\bfx}(\bfx,t)$, which is perpendicular to the edge $\nabla I(\bfx,t)$, is called \emph{normal flow}.
While \eqref{eq:supp:normal_flow_equation_1} is well-known in conventional (frame-based) computer vision, it is not directly applicable to event cameras because the latter do not directly provide the means to compute all terms required (the spatial and temporal derivatives of the brightness).
Next, we present how to compute the normal flow in the case of event cameras.

\subsubsection{Computation of Event-based Normal Flow.}
The output of an event camera is a stream of events, where each event $\boldsymbol{e}_{k}\doteq(\bfx_k,t_k,p_k)$ consists of the space-time coordinates $(\bfx_k,t_k)$ at which the intensity change of predefined size happened and the sign of the change (\ie, polarity $p_k\in\{+1,-1\}$). 
Following the definition in \cite{Benosman14tnnls}, we utilize the differential mapping function $\mathrm{\Sigma}_{e}: \Real^2 \to \Real$ that maps the pixel coordinate to the latest event's timestamp, i.e., $\bfx \mapsto t_\text{last}(\bfx)$. 
$\mathrm{\Sigma}_{e}$ is a 2D image, also called a time map; and it is often referred to as \emph{time surface} (TS) because when interpreting the image as an elevation map, a moving edge produces events and such points $(\bfx_k, t_\text{last}(\bfx_k)) \subset \Real^3$ approximately form a surface (ignoring the discrete pixel lattice) \cite{Benosman14tnnls,Lagorce17pami}.
It also goes by the name of \emph{Surface of Active Events} (SAE).\\

\noindent
\emph{Lemma.}
\emph{Given the mapping function $\mathrm{\Sigma}_e$ at pixel coordinate $\bfx$ and its finite neighbourhood, the direction of the first-order partial derivative ${\nabla}\mathrm{\Sigma}_e(\bfx) = \frac{\partial \mathrm{\Sigma}_e}{\partial \bfx} (\bfx)$ is identical to that of the normal flow at $\bfx$}.\\

\noindent
\emph{Proof.} 
The directional derivative ${\nabla }_{\bfd}\mathrm{\Sigma}_e(\bfx)$ can be defined by the dot product of direction $\bfd$ and ${\nabla}\mathrm{\Sigma}_e(\bfx)$, as ${\nabla }_{\bfd}\mathrm{\Sigma}_e(\bfx) \doteq \bfd \cdot {\nabla}\mathrm{\Sigma}_e(\bfx)$, whose maximum is achieved when $\bfd_{\max} = \frac{ {\nabla}\mathrm{\Sigma}_e(\bfx)}{\left \| {\nabla}\mathrm{\Sigma}_e(\bfx) \right \| }$.
The direction $\bfd_{\max}$ is proved to be perpendicular to the level set $\mathcal{S} \doteq \{ \bfx | \mathrm{\Sigma}_e(\bfx) = t \}$ which corresponds to the latent edge pixels at time $t$.
Hence, direction $\bfd_{\max}$ is in parallel to the direction of local image gradient, namely the normal flow's direction.
$\hfill\square$\\

The lemma demonstrates how to determine the direction of normal flow, but there is still a unknown scale factor between the magnitude of $\bfd_{\max}$ and that of the normal flow.\\

\noindent
\emph{Proposition.}
\emph{Given the mapping function $\mathrm{\Sigma}_e$ at pixel coordinate $\bfx$ and its finite neighbourhood, and the first-order partial derivative ${\nabla}\mathrm{\Sigma}_e(\bfx)$, normal flow $\bfn(\bfx)$ at pixel $\bfx$ can be calculated by}
\begin{equation}
    \label{eq:supp:normal_flow_computtion_from_events}
    \bfn(\bfx) = \frac{{\nabla}\mathrm{\Sigma}_e(\bfx)}{\left \|  {\nabla }\mathrm{\Sigma}_e(\bfx)\right \|^2}.
\end{equation}

\noindent
\emph{Proof.}
Direction $\bfd_{\max}$ owns the algebra meaning: the maximum time increase magnitude $\mathrm{\Delta}{t}_{\max} = \bfd_\text{max}\cdot\nabla\mathrm{\Sigma}_{e}(\bfx) = \left \| {\nabla}\mathrm{\Sigma}_e(\bfx)\right \|$ given a unit pixel displacement ($\|\bfd_{\max}\|=1$). 
Since the magnitude of the normal flow is equal to the pixel displacement per unit time, dividing $\bfd_{\max}$ by $\mathrm{\Delta}{t}_{\max}$ is the normal flow, i.e., $\bfn(\bfx)=\frac{\bfd_{\max}}{\mathrm{\Delta}{t}_{\max}}=\frac{\nabla\mathrm{\Sigma}_{e}(\bfx)}{\|\nabla\mathrm{\Sigma}_{e}(\bfx)\|^2}$.
$\hfill\square$\\

\begin{figure}[t]
\centering
\includegraphics[width=\textwidth]{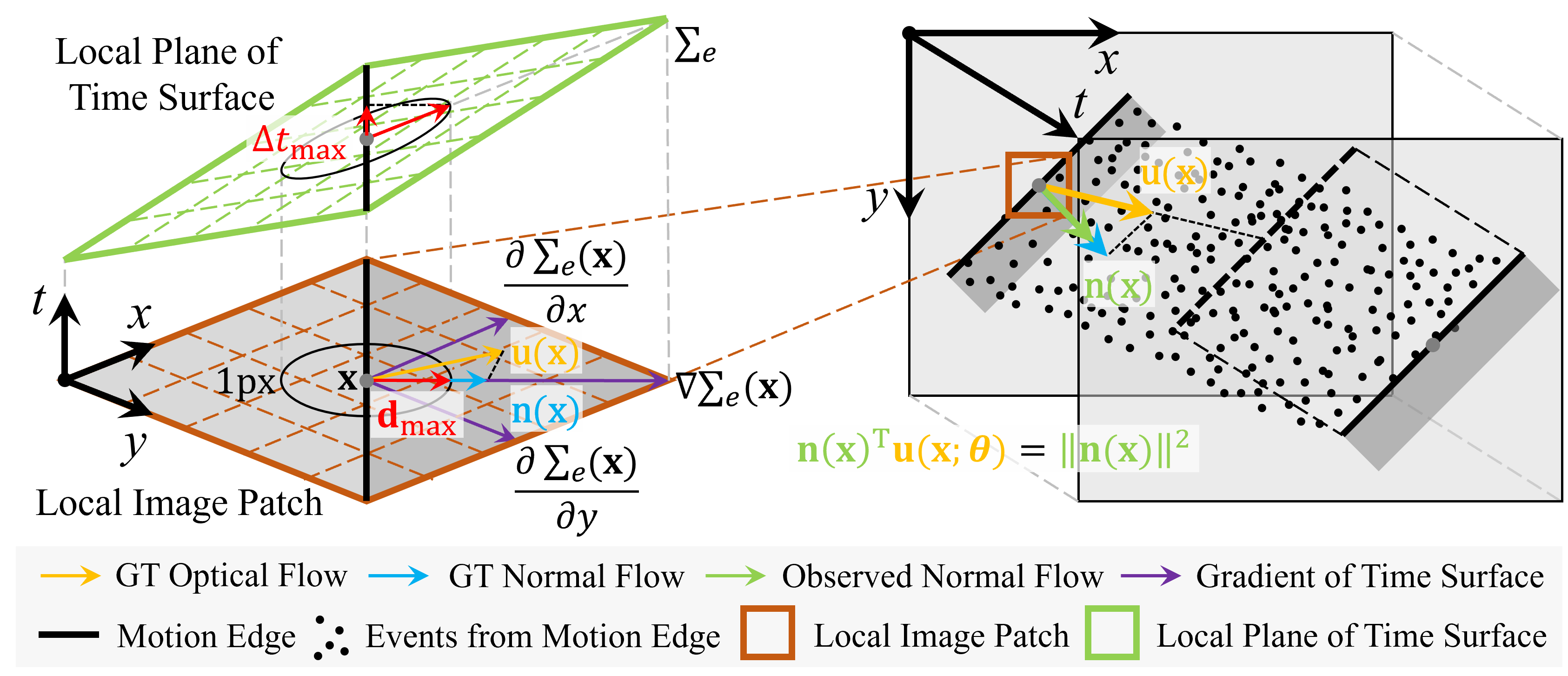}
\caption{
Illustration of event-based normal flow, explained on the spatio-temporal profile induced by a moving edge.
Given a pixel coordinate $\bfx$, when the increment $\mathrm{\Delta}\bfx$'s direction $\mathbf{d}$ equals $\bfd_{\max}$ ($\bfd_{\max}$ is the one that maximizes the directional derivative $\nabla_{\bfd}\mathrm{\Sigma}_{e}(\bfx)=\bfd\cdot\nabla\mathrm{\Sigma}_{e}(\bfx)$, i.e., \ $\bfd_{\max}=\frac{\nabla\mathrm{\Sigma}_{e}(\bfx)}{\|\nabla\mathrm{\Sigma}_{e}(\bfx)\|}$), it aligns with the direction of the time surface gradient $\nabla\mathrm{\Sigma}_{e}(\bfx)$, thereby determining the direction of the normal flow $\bfn(\bfx)$.
Since the time increment (lifetime) corresponding to $\bfd_{\max}$ is $\mathrm{\Delta}{t}_{\max}=\bfd_{\max}\cdot\nabla\mathrm{\Sigma}_{e}(\bfx)=\|\nabla\mathrm{\Sigma}_{e}(\bfx)\|$, the normal flow $\bfn(\bfx)=\frac{\bfd_{\max}}{\mathrm{\Delta}{t}_{\max}}=\frac{\nabla\mathrm{\Sigma}_{e}(\bfx)}{\|\nabla\mathrm{\Sigma}_{e}(\bfx)\|^2}$.
}
\label{fig:normal_flow_constraint}
\end{figure}

The above normal flow calculation requires the smooth and differentiable mapping function, we can approximate its partial derivative operator with the modern image processing techniques such as \emph{Sobel filter} or \emph{local plane fitting}. 
All above-mentioned geometric elements are illustrated in Fig.~\ref{fig:normal_flow_constraint}.
Based on this, we present the normal flow constraint $\bfn(\bfx)^{\top}\bfu(\bfx)=\|\bfn(\bfx)\|^2$ for robust estimation of motion and structure parameters in Sec. 4 of our paper.

\section{More About Differential Homography}
\label{sec:method_supp}
In this section, a detailed explanation to Sec.~4 of our paper is added.
In particular, we first discuss the way to recover the true differential homography from our linear solution (Sec. \ref{subsec:recover_true_differential_homography_matrix_from_linear_solver}).
Then, we disclose how to retrieve the motion and scene parameters by decomposing differential homography (Sec.~\ref{subsec:Decompose Differential Homography To Get The Motion Parameters}).

\subsection{Recover the True Differential Homography from Our Linear Solution}
\label{subsec:recover_true_differential_homography_matrix_from_linear_solver}
As is well-known \cite{Ma04book,Differential_Homo_ECCV20}, Eq.~\eqref{eq:supp:differ_homo} can only recover $\mathbf{H}_L$ = $\mathbf{H}_d$ +$\epsilon \mathbf{I}$ with an unknown scale $\epsilon$ since $\mathbf{H}_d$ has a one-dimensional null space.
So our linear solver faces this problem too, here we could recover the true differential homography matrix as follows: 
after solving for $\mathbf{H}_L$, let $\mathbf{M}_L = \mathbf{H}_L+ \mathbf{H}_L^{\top}$, 
then the second largest eigenvalue of $\mathbf{M}_L$ equals $2\epsilon$, which enabling recovery of differential homography matrix by $\mathbf{H}_d = \mathbf{H}_L - \epsilon\mathbf{I}$.

\noindent \emph{Proof.} Let 
\begin{equation}
\begin{aligned}
    \mathbf{M}_L &= - \left(\frac{\boldsymbol{\nu}}{d}\mathbf{n}^{\top}+\mathbf{n}{\frac{\boldsymbol{\nu}}{d}}^{\top}\right) + 2 \epsilon \mathbf{I} \\
    &= -\mathbf{M} + 2 \epsilon \mathbf{I}.
\end{aligned}
\end{equation}
We define the eigenvalue of $\mathbf{M}$ as $\lambda_{\max}, \lambda_{\text{mid}}, \lambda_{\min}$ with the decreasing order. Obviously $\mathbf{M}$ is a symmetric matrix, so according to the Rayleigh quotient theorem, for any vector $\{ \mathbf{x} | \mathbf{x} \in \Real^3,  \left \| \mathbf{x} \right \|  = 1 \}$ we have
\begin{equation}
    \lambda_{\min}\mathbf{x}^{\top}\mathbf{x} \le \mathbf{x}^{\top} \mathbf{M} \mathbf{x} \le \lambda_{\max}\mathbf{x}^{\top}\mathbf{x}.
\end{equation}
For simplicity, we omit $d$ and define a new variable $\mathbf{v} = {\boldsymbol{\nu}}/{d}$.
Taking a vector $\mathbf{x}$ perpendicular to ($\mathbf{v}+\mathbf{n}$), and not perpendicular to $\mathbf{v}$ and $\mathbf{n}$ at the same time, we can prove that 
\begin{equation}
\begin{aligned}
\lambda_{\min}\mathbf{x}^{\top}\mathbf{x}   & \le \mathbf{x}^{\top} \left(\mathbf{n}\mathbf{v}^{\top}+\mathbf{v}\mathbf{n}^{\top} \right)\mathbf{x}\\
 & =\mathbf{x}^{\top} ((\mathbf{v}+\mathbf{n})(\mathbf{v}+\mathbf{n})^{\top}-(\mathbf{v}\mathbf{v}^{\top}+\mathbf{n}\mathbf{n}^{\top} ))\mathbf{x}\\
 & = -\mathbf{x}^{\top} (\mathbf{v}\mathbf{v}^{\top}+\mathbf{n}\mathbf{n}^{\top} )\mathbf{x}\\
 & = -[(\mathbf{x}^{\top} \mathbf{v})^2+(\mathbf{x}^{\top}\mathbf{n})^2 ] <  0.
\end{aligned}
\end{equation}
And we know that the inner product of the vector itself $\mathbf{x}^{\top}\mathbf{x}$ is always positive. We can then conclude that $\lambda_{\min}<0$.

Similarly, for another $\mathbf{x}$ perpendicular to ($\mathbf{v}-\mathbf{n}$), and not perpendicular to $\mathbf{v}$ and $\mathbf{n}$ at the same time we can prove that
\begin{equation}
\begin{aligned}
\lambda_{\max}\mathbf{x}^{\top}\mathbf{x}   & \ge \mathbf{x}^{\top} (\mathbf{n}\mathbf{v}^{\top}+\mathbf{v}\mathbf{n}^{\top}   )\mathbf{x}\\
 & = \mathbf{x}^{\top}((\mathbf{v}\mathbf{v}^{\top}+\mathbf{n}\mathbf{n}^{\top} )- (\mathbf{v}-\mathbf{n})(\mathbf{v}-\mathbf{n})^{\top} )\mathbf{x}\\
 & = \mathbf{x}^{\top} (\mathbf{v}\mathbf{v}^{\top}+\mathbf{n}\mathbf{n}^{\top} )\mathbf{x}\\
 & = [(\mathbf{x}^{\top} \mathbf{v})^2+(\mathbf{x}^{\top}\mathbf{n})^2 ] >  0.
\end{aligned}
\end{equation}

Similarly, we can prove that $\lambda_{\max}>0$.

For any rank deficient square matrix, its determinant must equal to zero and there are at least one eigenvalue equals to zero. The rank of the $\mathbf{M}$ is less than two since it contains two outer product of vectors
\begin{equation}
    \text{rank}(\mathbf{M}) 
    =\text{rank}\left(\frac{\boldsymbol{\nu}}{d}\mathbf{n}^{\top}+\mathbf{n}\frac{\boldsymbol{\nu}}{d}^{\top}\right)\le 2.
\end{equation}
Besides, $\lambda_{\max}>0$ and $\lambda_{\min}<0$, we can come to the conclusion that 
\begin{equation}
    \lambda_{\text{mid}} = 0.
\end{equation}
Consequently the second largest value of $\mathbf{M}_L = -\mathbf{M} + 2 \epsilon \mathbf{I}$ equals to $2\epsilon$. 
\hfill $\square$

\subsection{Retrieving Motion and Structure Parameters by Decomposing Differential Homography}
\label{subsec:Decompose Differential Homography To Get The Motion Parameters}
After recovering the real differential homography matrix $\mathbf{H}_d$, this section introduces the decomposition method to get the motion and structure parameters. 

Firstly, the eigen-decomposition of $\mathbf{M}$ can be written as:
\begin{equation}
\label{eq:supp:eigendecom}
\mathbf{M}=\mathbf{Q}\boldsymbol{\Lambda}\mathbf{Q}^{\top}.
\end{equation}
where $\mathbf{Q} = \{\mathbf{q}_{\max}, \mathbf{q}_\text{mid}, \mathbf{q}_{\min} \}$ is the orthogonal matrix which contains all eigenvectors of $\mathbf{M}$ and $\text{diag}(\boldsymbol{\Lambda}) = \{\lambda_{\max}, \lambda_{\text{mid}}, \lambda_{\min} \}$ which collects all eigenvalues in the diagonal elements.

We then can define two auxiliary vectors $\mathbf{j}$ and $\mathbf{k}$ 
\begin{equation}
\label{eq:supp:jk}
    \begin{aligned}
        \mathbf{j} &= \sqrt{\frac{\lambda_{\max}}{2}}\mathbf{q}_{\max} + \sqrt{\frac{-\lambda_{\min}}{2}}\mathbf{q}_{\min}, \\
        \mathbf{k} &= \sqrt{\frac{\lambda_{\max}}{2}}\mathbf{q}_{\max} - \sqrt{\frac{-\lambda_{\min}}{2}}\mathbf{q}_{\min},
    \end{aligned}
\end{equation}
such that 
\begin{equation}
\label{eq:supp:m_jk}
\mathbf{M} = \mathbf{j}\mathbf{k}^{\top} + \mathbf{k}\mathbf{j}^{\top}.
\end{equation}
\noindent \emph{Proof.}

In Sec. \ref{subsec:recover_true_differential_homography_matrix_from_linear_solver}, we already concluded that the egienvalues of $\mathbf{M}$ have following properties:
\begin{equation}
\lambda_{\max} > \lambda_{\text{mid}} = 0 > \lambda_{\min}.
\end{equation}
Then, we can substitute this result into Eq.~(\ref{eq:supp:eigendecom}):
\begin{equation}
\begin{aligned}
        \mathbf{M} &= \begin{bmatrix}
        \mathbf{q}_{\max}   &\mathbf{q}_\text{mid}  &\mathbf{q}_{\min}
    \end{bmatrix}
    \begin{bmatrix}
        \lambda_{\max}  &   &\\
                            &0  &\\
                            &   &\lambda_{\min}
    \end{bmatrix}
    \begin{bmatrix}
        {\mathbf{q}_{\max}}^{\top}\\
        {\mathbf{q}_\text{mid}}^{\top}\\
        {\mathbf{q}_{\min}}^{\top}
    \end{bmatrix}\\
    &= \lambda_{\max} \bfq_{\max}{\bfq_{\max}}^{\top} + \lambda_{\min}\bfq_{\min}{\bfq_{\min}}^{\top} \\
    &= \frac{\lambda_{\max}}{2} \bfq_{\max}{\bfq_{\max}}^{\top} + \frac{\lambda_{\max}}{2} \bfq_{\max}{\bfq_{\max}}^{\top} +\\
    &~~~~\frac{\lambda_{\min}}{2}\bfq_{\min}{\bfq_{\min}}^{\top} + \frac{\lambda_{\min}}{2}\bfq_{\min}{\bfq_{\min}}^{\top}\\
    &= (\frac{\lambda_{\max}}{2} \bfq_{\max}{\bfq_{\max}}^{\top} + \sqrt{\frac{-\lambda_{\min}\lambda_{\max}}{4}}\bfq_{\min}{\bfq_{\max}}^{\top}\\ 
    &~~~~ - \sqrt{\frac{-\lambda_{\min}\lambda_{\max}}{4}}\bfq_{\max}{\bfq_{\min}}^{\top} - \frac{-\lambda_{\min}}{2}\bfq_{\min}{\bfq_{\min}}^{\top})\\
    &~~~~ +
    (\frac{\lambda_{\max}}{2} \bfq_{\max}{\bfq_{\max}}^{\top} - \sqrt{\frac{-\lambda_{\min}\lambda_{\max}}{4}}\bfq_{\min}{\bfq_{\max}}^{\top}\\
    &~~~~ + \sqrt{\frac{-\lambda_{\min}\lambda_{\max}}{4}}\bfq_{\max}{\bfq_{\min}}^{\top} - \frac{-\lambda_{\min}}{2}\bfq_{\min}{\bfq_{\min}}^{\top}).
\end{aligned}
\end{equation}
It is easy to prove
\begin{equation}
\begin{aligned}
\bfj\bfk^{\top}=
    & \left(\frac{\lambda_{\max}}{2} \bfq_{\max}{\bfq_{\max}}^{\top} + \sqrt{\frac{-\lambda_{\min}\lambda_{\max}}{4}}\bfq_{\min}{\bfq_{\max}}^{\top} \right.\\
    &\quad \left. - \sqrt{\frac{-\lambda_{\min}\lambda_{\max}}{4}}\bfq_{\max}{\bfq_{\min}}^{\top} - \frac{-\lambda_{\min}}{2}\bfq_\text{min}{\bfq_{\min}}^{\top}\right),\\
    \bfk\bfj^{\top} =
    &\left( \frac{\lambda_{\max}}{2} \bfq_{\max}{\bfq_{\max}}^{\top} - \sqrt{\frac{-\lambda_{\min}\lambda_{\max}}{4}}\bfq_{\min}{\bfq_{\max}}^{\top} \right.\\
    &\quad \left. + \sqrt{\frac{-\lambda_{\min}\lambda_{\max}}{4}}\bfq_{\max}{\bfq_{\min}}^{\top} - \frac{-\lambda_{\min}}{2}\bfq_{\min}{\bfq_{\min}}^{\top}\right).
\end{aligned}
\end{equation}
We then can conclude that
$\bfM = \bfj\bfk^{\top} + \bfk\bfj^{\top}$.
$\hfill\square$

Since $\mathbf{M}$ is a symmetric matrix, we cannot directly distinguish ${\boldsymbol{\nu}}/{d}$ and $\mathbf{n}$ from each other. Note that $\mathbf{n}$ is a normal vector which has a unique norm. We can thus decompose all candidates as
\begin{equation}
\label{eq:supp:vn}
\left\{
    \begin{array}{ll}
    \frac1{d}\linvel = \bfj \|\bfk\|, \quad \bfn = \bfk / \|\bfk\|, &\quad \angvel = -(\bfH_d+ \bfj \bfk^{\top})^{\vee}  \\[1ex]
    \frac1{d}\linvel = \bfk \|\bfj\|, \quad \bfn = \bfj / \|\bfj\|, &\quad \angvel = -(\bfH_d+ \bfk \bfj^{\top})^{\vee}, 
    \end{array}
\right.
\end{equation}
where the $(.)^{\vee}$ denotes the inverse operator of $(.)_{\times}$, which converts a skew-symmetric matrix into a vector.

Above all, the decomposition of the differential homography matrix results in two sets of possible solutions that are indistinguishable in the absence of prior scene or motion information. 
However, when integrated with data from other sensors, this decomposition remains valuable, enabling the estimation of otherwise unknown motion and scene parameters.

\section{Dataset and Evaluation Metrics}
\label{sec:dataset_supp}
To supplement Sec. 5 of our paper, we detail our dataset as well as the evaluation metrics used in the experiments.
\subsection{Our Dataset}

\subsubsection{Real Data.}
\begin{figure}[htbp]
    \centering
    \begin{subfigure}{0.48\linewidth}
    \includegraphics[width=0.48\textwidth]{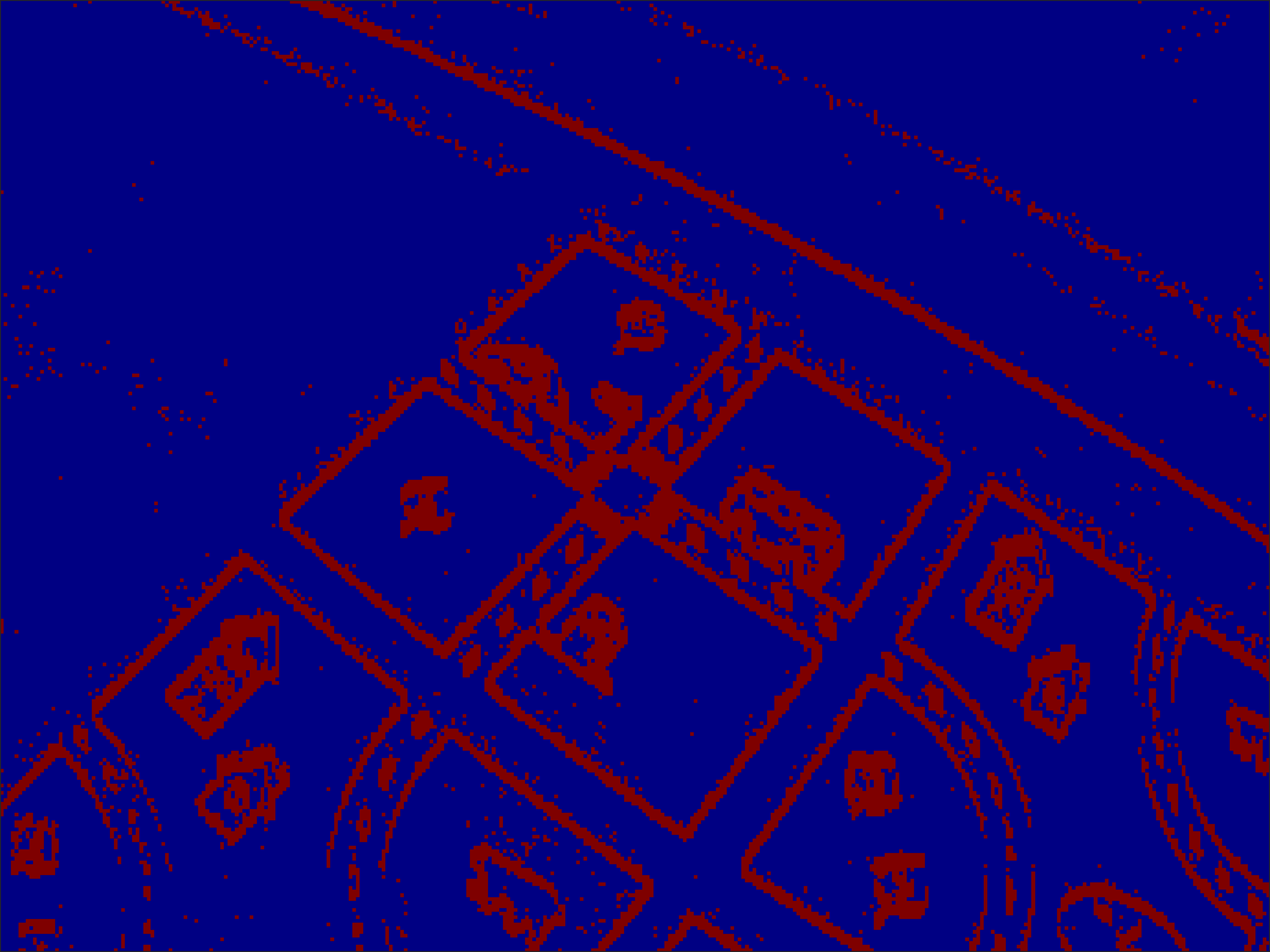}
    \includegraphics[width=0.48\textwidth]{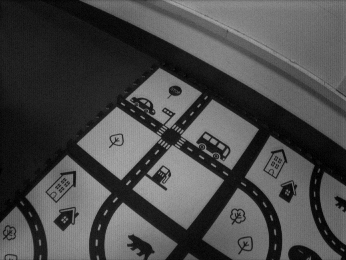}
    \caption{\emph{ground\_rotation}}
    \label{subfig:eg_ground_rotation}
    \end{subfigure}
    \;
    \begin{subfigure}{0.48\linewidth}
    \includegraphics[width=0.48\textwidth]{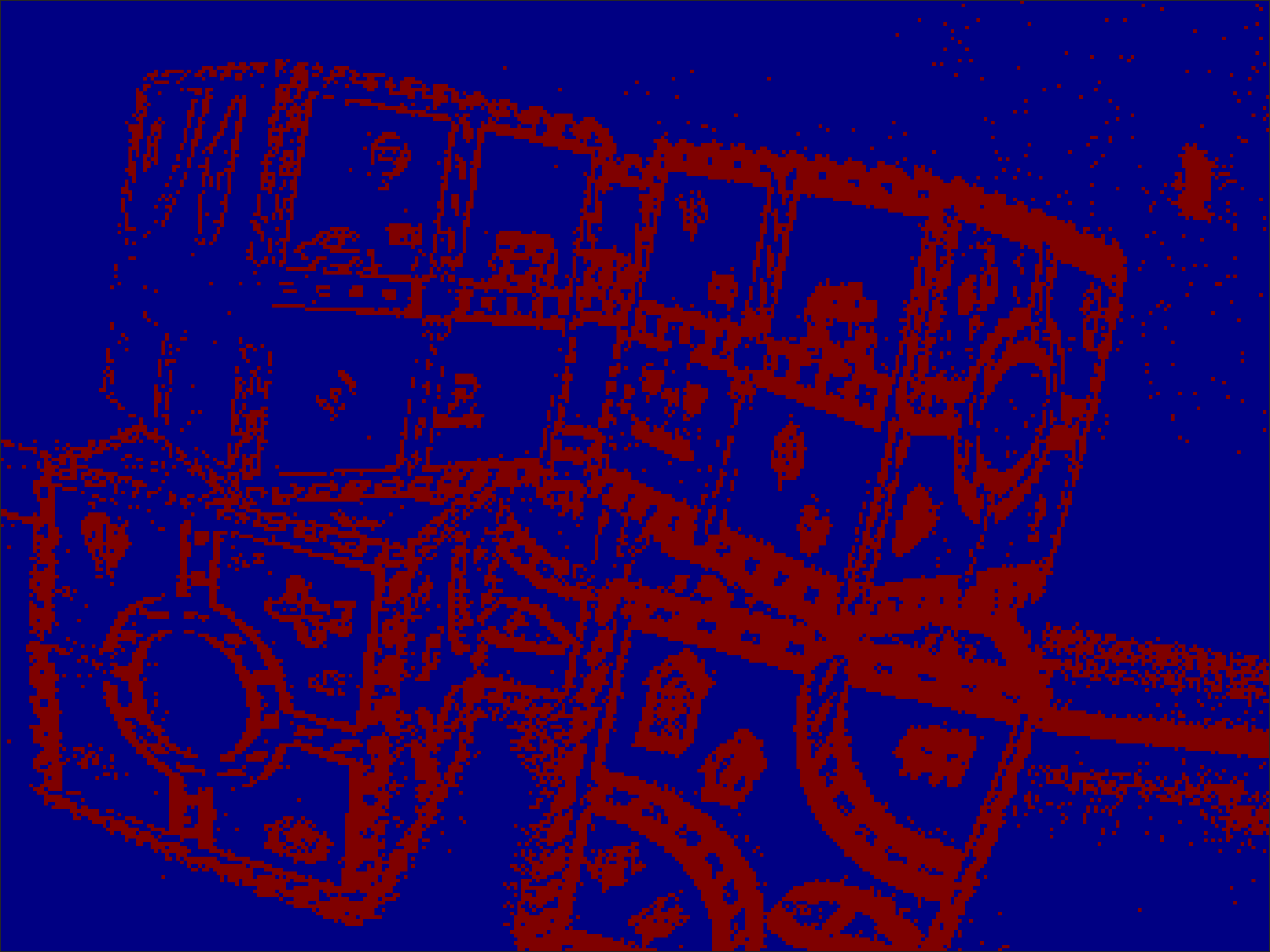}
    \includegraphics[width=0.48\textwidth]{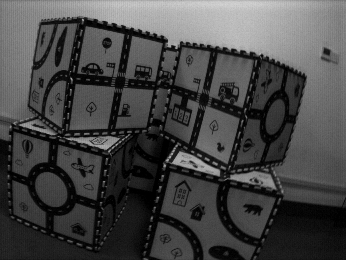}
    \caption{\emph{boxes\_rotation}}
    \label{subfig:eg_boxes_rotation}
    \end{subfigure}
    \caption{Event data (represented with a naive accumulation) and corresponding frames from our data collected using a DAVIS-346 event camera.}
    \label{fig:davis346_eg}
\end{figure}

\begin{figure}[htbp]
    \centering
    \begin{subfigure}{\linewidth}
    \centering
    \includegraphics[width=.85\linewidth]{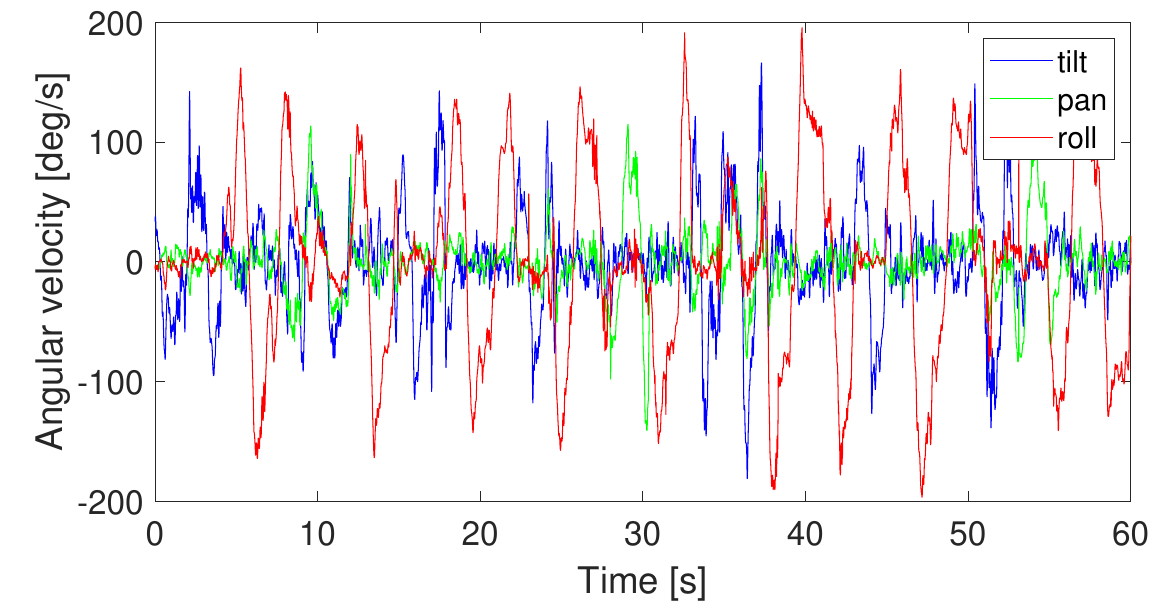}
    \caption{\emph{ground\_rotation}}
    \label{subfig:gt_ground_rotation}
    \end{subfigure}
    \begin{subfigure}{\linewidth}
    \centering
    \includegraphics[width=.85\linewidth]{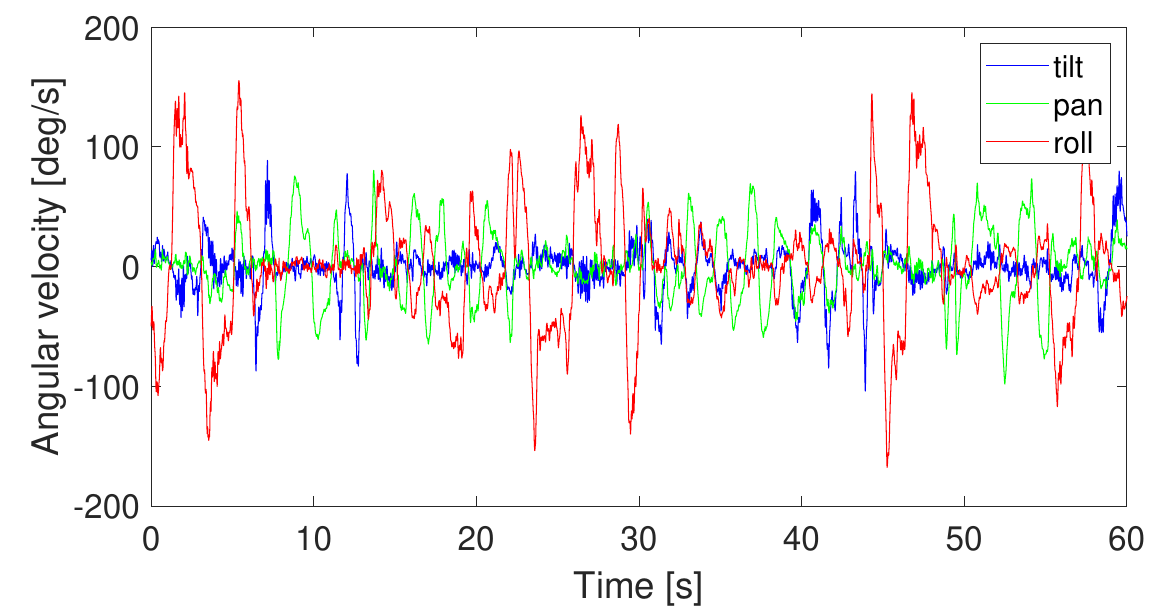}
    \caption{\emph{boxes\_rotation}}
    \label{subfig:gt_boxes_rotation}
    \end{subfigure}
    \caption{Angular velocity measurements from an IMU, used as ground truth.}
    \label{fig:davis346_gt}
\end{figure}

For our study on rotational motion estimation, we initially performed experiments using the ECD dataset\cite{Mueggler17ijrr}, which was captured using an event camera\cite{Brandli14ssc} with a relatively low spatial resolution (a DAVIS240C with 240$\times$180 px). 
To assess our proposed algorithm with more modern sensors, we recorded two similar sequences, \emph{ground\_rotation} and \emph{boxed\_rotation}, using an event camera (iniVation DAVIS346) with a higher spatial resolution (346$\times$260 px). 
Additionally, we employ an Xsens Mti-630 IMU to provide ground-truth angular velocity. 
Fig.~\ref{fig:davis346_eg} offers an insight into the recorded data. 
Meanwhile, Fig.~\ref{fig:davis346_gt} depicts the ground-truth angular velocity information, featuring high dynamics of the event camera in each sequence.

\subsubsection{Synthetic Data.}
\begin{figure}[htbp]
    \centering
    \begin{subfigure}{0.48\linewidth}
    \includegraphics[width=0.49\textwidth]{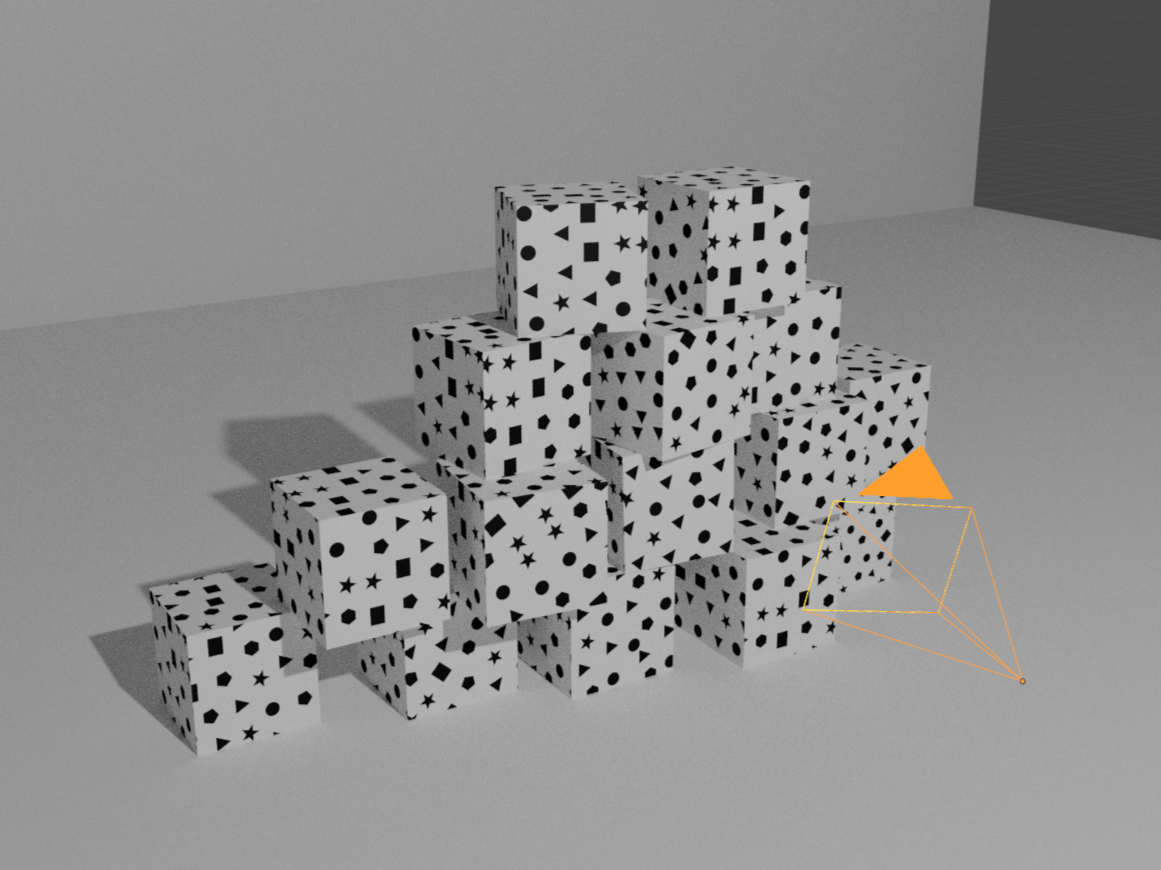}
    \includegraphics[width=0.49\textwidth]{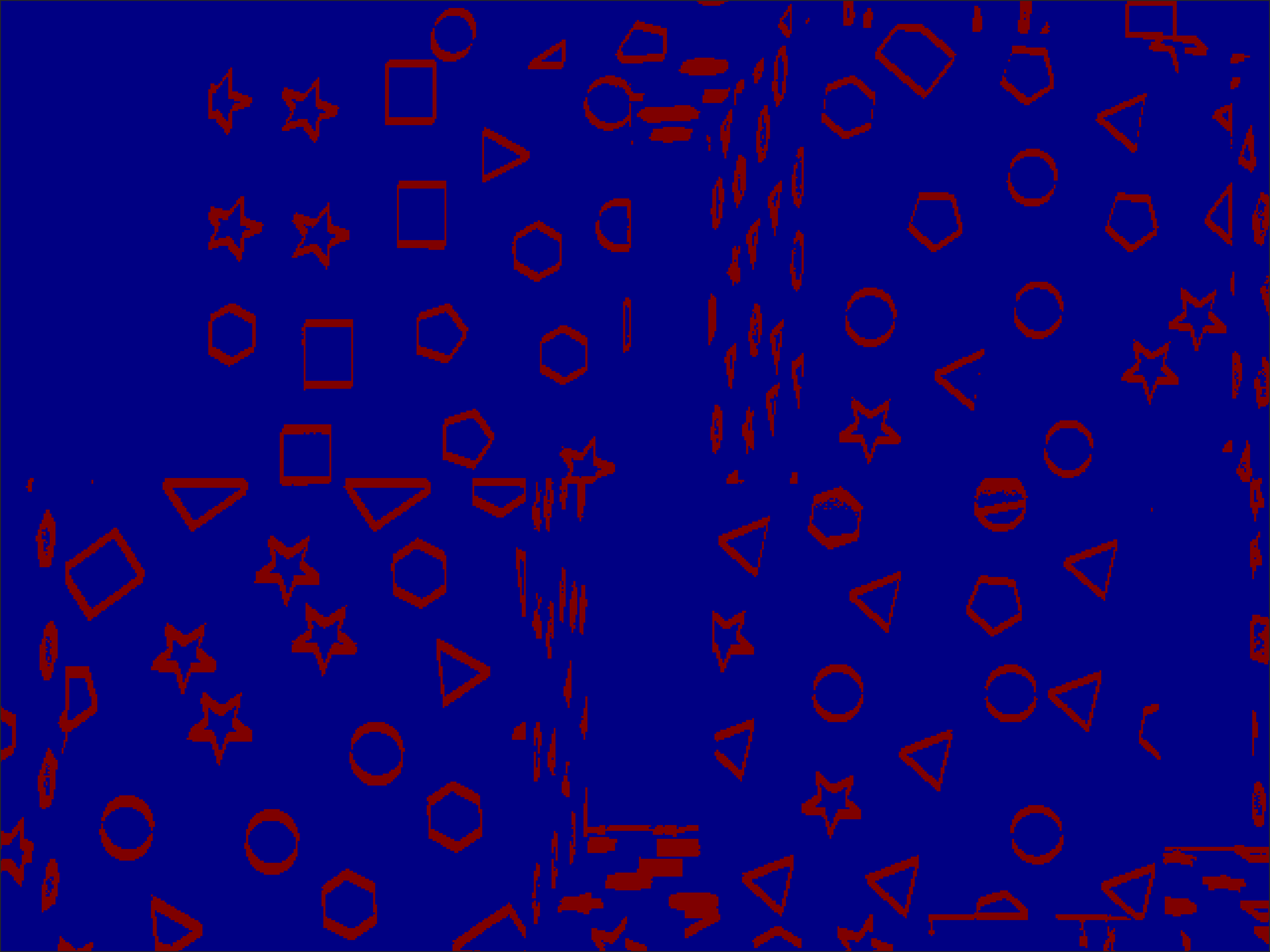}
    \caption{\emph{cubes\_rotation}}
    \label{subfig:cubes_blender}
    \end{subfigure}
    \;
    \begin{subfigure}{0.48\linewidth}
    \includegraphics[width=0.49\textwidth]{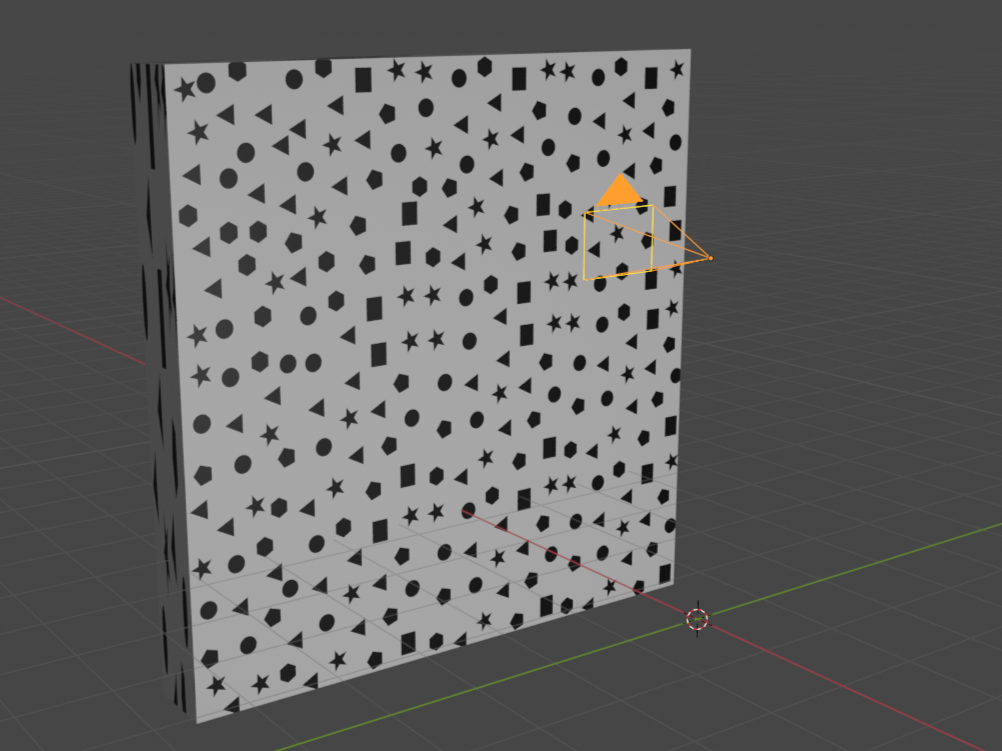}
        \includegraphics[width=0.49\textwidth]{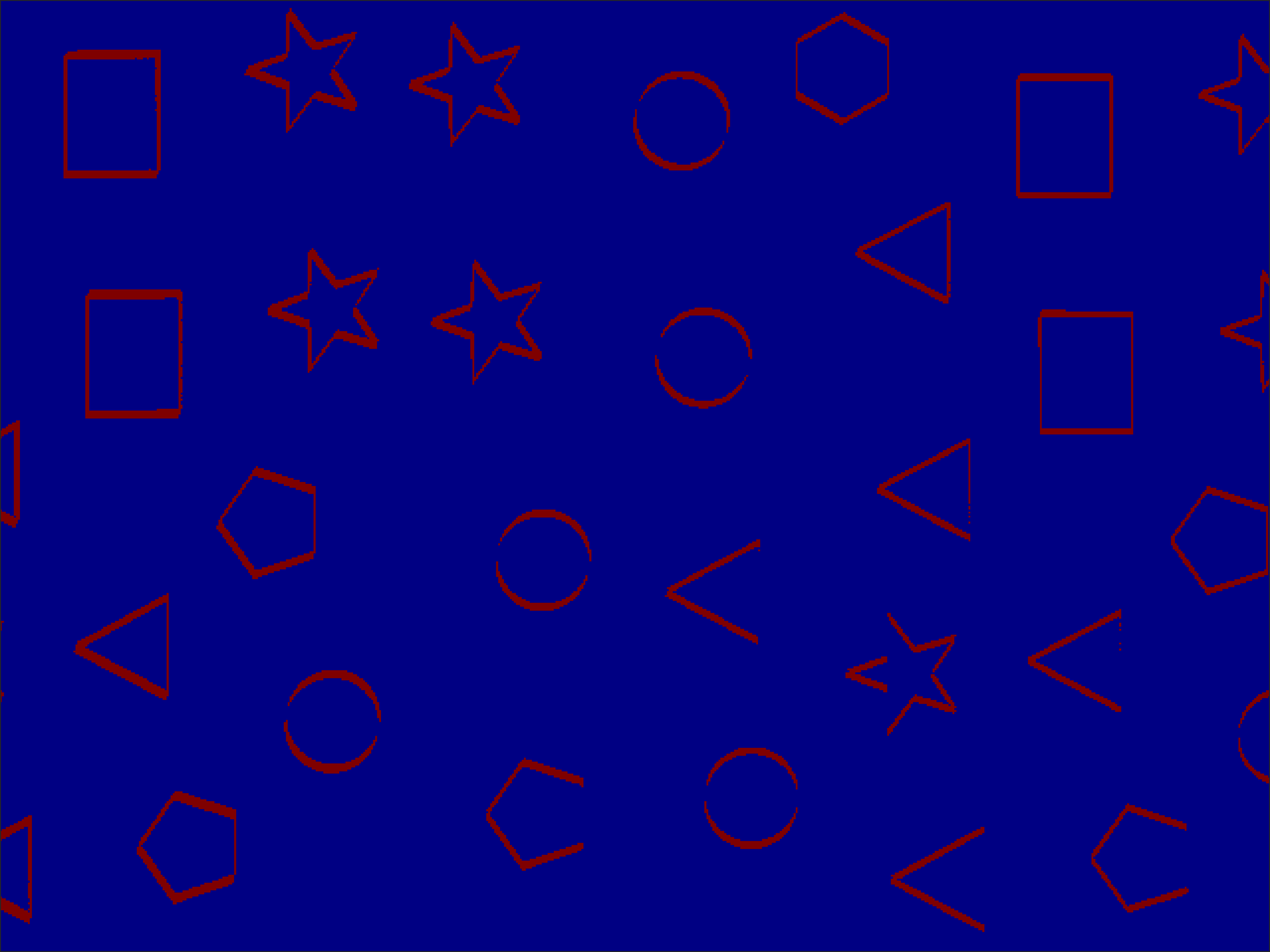}
\caption{\emph{patterns\_rotation}}
    \label{subfig:oneWall_blender}
    \end{subfigure}
    \caption{An event camera (orange) capturing different sequences in simulated scenes.}
    \label{fig:blender}
\end{figure}

In the paper, we conduct a comprehensive set of experiments using a synthetic dataset created with a simulator \cite{Rebecq18corl}. 
This dataset comprises sequences, including \emph{two\_wall\_translation}, \emph{patterns\_rotation}, \emph{cubes\_rotation}, \emph{patterns\_6dof}, and \emph{cubes\_6dof}, synthesized using an event camera with a spatial resolution of 640$\times$480 px. 
Additionally, the simulator provides us with ground truth data of the optical flow, depth and camera trajectory for each sequence. 
All sequences are recorded over a time span of 0.5 seconds under different motion patterns. 
Specifically, sequences \emph{patterns\_6dof} and \emph{cubes\_6dof} feature a combined motion pattern of a linear translation plus a rotation around a certain axis.
In contrast, sequences \emph{patterns\_rotation} and \emph{cubes\_rotation} involve pure rotation at a constant angular velocity, while sequence \emph{two\_wall\_translation} exhibits a pure translation.
\Cref{fig:blender} shows the simulated scenes and the pose of the camera (represented by an orange wireframe pyramid).

\subsection{Evaluation Metrics}
In the quantitative evaluation, we utilize the average angular velocity error ($e_w$) and root mean square error (RMSE) to measure the angular velocity estimation error.
The latter is defined as 
\begin{equation}
    {\rm RMSE}_\omega = \sqrt{\frac{1}{3m}\sum_{i=1}^{m} \left(e_{\omega_xi}^2+e_{\omega_yi}^2+e_{\omega_zi}^2\right)}.
\end{equation}
For linear velocity evaluation, we also evaluate error by means of the average ($e_v$) and the RMSE, 
with
\begin{equation}
    {\rm RMSE}_v = \sqrt{\frac{1}{3m}\sum_{i=1}^{m} \left(e_{v_xi}^2+e_{v_yi}^2+e_{v_zi}^2\right)}.
\end{equation}

As for the evaluation metric for the differential homography, we assess the errors in the estimated differential homography matrix by means of the Frobenius norm of the difference with respect to the ground truth, rather than directly comparing motion and structure parameters.
This is because it is non-trivial to determine the correct one from the resulting two-set motion and structure parameters (see \ref{subsec:Decompose Differential Homography To Get The Motion Parameters}), and it is out of the scope of this paper.

\section{Implementation Details}
\label{sec:implementation_supp}
This section provides implementation details to supplement Sec.~5 of the paper.

\subsubsection{Normal Flow Extraction.}
Since our minimal solver is built on normal flow constraint, the pre-stage normal flow calculation becomes crucial. 
Here, we directly calculate gradients from raw event data by local plane fitting \cite{Benosman14tnnls,Mueggler15icra}.
To further refine these calculations, RANSAC is applied. 
Ultimately, we employ \eqref{eq:supp:normal_flow_computtion_from_events} to accurately determine the partial normal flow.

\subsubsection{Linear Solver.}
From the event batch, we extracted numerous partial normal flow constraints, which enabled us to construct an overdetermined system of equations.
However, the estimated normal flow is still subject to noise due to the limitations of existing methods.
To address this, we employed RANSAC for robust sampling.

\subsubsection{Hyper-Parameter Settings.}
Parameters including window size and threshold can have a crucial influence on the performance of each solver, which therefore needs clarification.
Here, we have configured the spatial and temporal window sizes for normal flow extraction at $7\text{px} \times 7\text{px} \times 0.04\text{s}$, respectively. 
Additionally, the RANSAC threshold of plane fitting is set to $10^{-5}$, while the RANSAC threshold of linear solver is established at $10^{-4}$.

\section{Additional Results}
\label{sec:extensive_experiments_supp}
To further investigate and discuss the performance of our solver, we conduct some extensive experiments to supplement Sec. 5 of the paper.
\begin{figure*}[htbp]
\centering
\begin{subfigure}{0.32\linewidth}
\includegraphics[width=\textwidth]{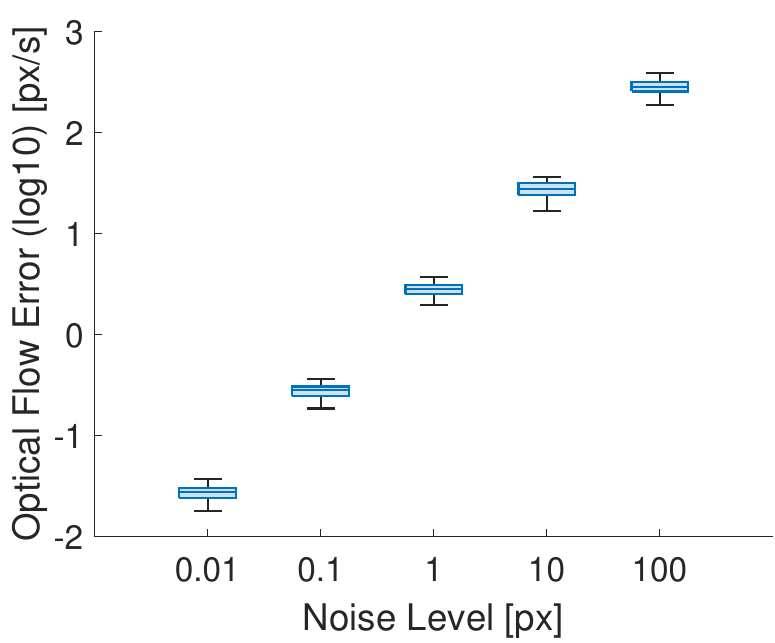}
\caption{Optical Flow}
\label{fig:numerical oflow}
\end{subfigure}
\begin{subfigure}{0.32\linewidth}
\includegraphics[width=\textwidth]{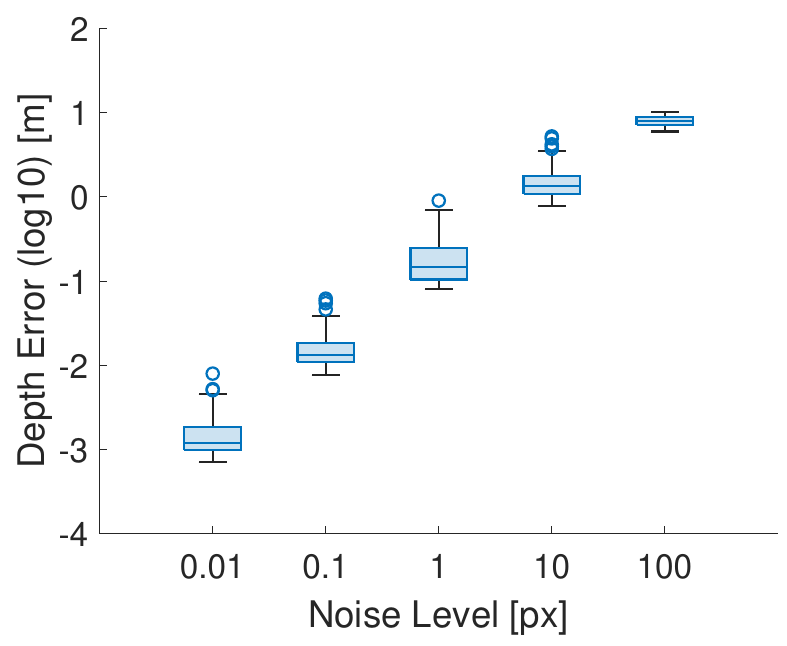}
\caption{Depth}
\label{fig:numerical depth}
\end{subfigure}
\begin{subfigure}{0.32\linewidth}
\includegraphics[width=\textwidth]{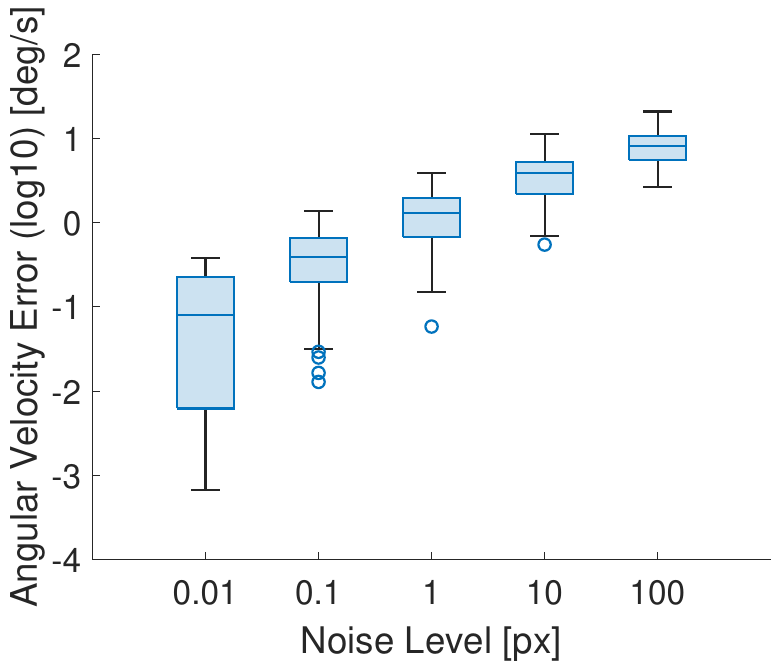}
\caption{Rotational Motion}
\label{fig:numerical rotaion}
\end{subfigure}
\begin{subfigure}{0.32\linewidth}
\includegraphics[width=\textwidth]{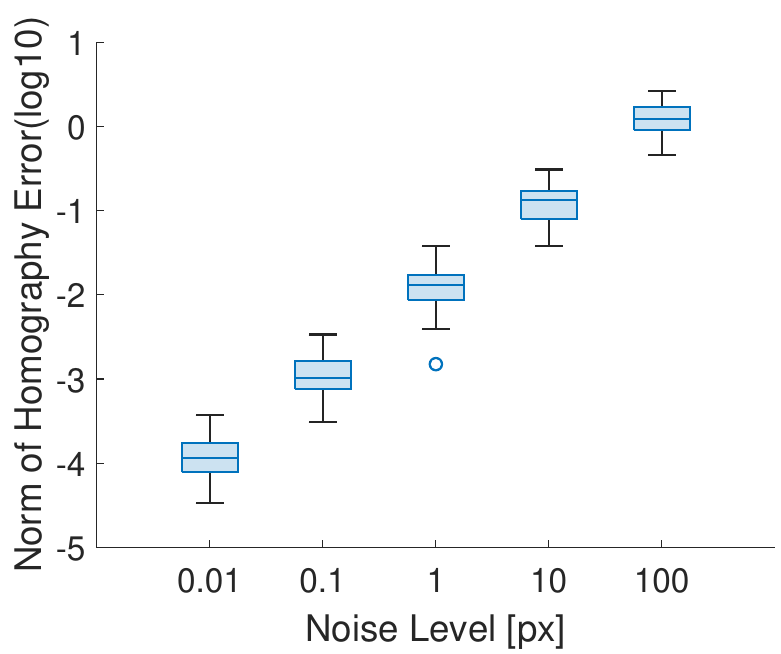}
\caption{Differential Homography}
\label{fig:numerical homo}
\end{subfigure}
\begin{subfigure}{0.32\linewidth}
\includegraphics[width=\textwidth]{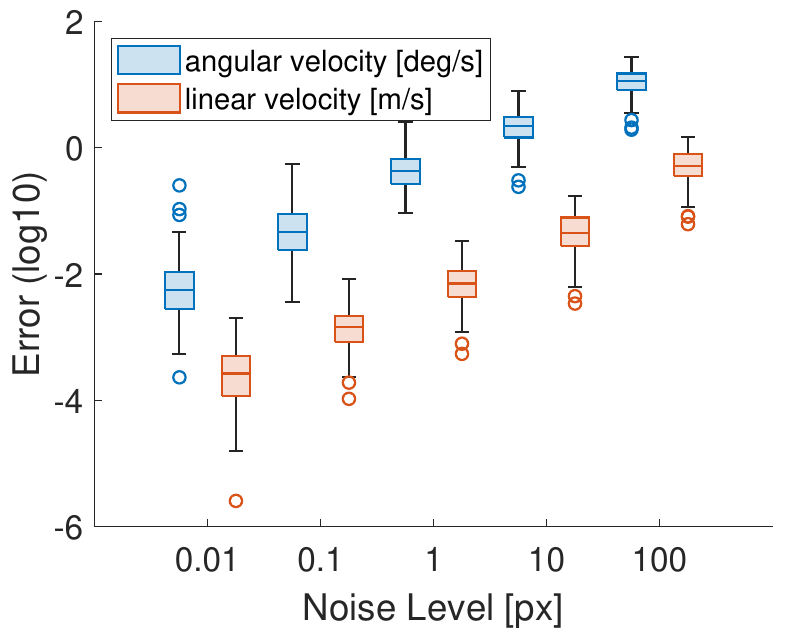}
\caption{6-DoF Motion}
\label{fig:numerical tracking}
\end{subfigure}
\caption{Numerical stability analysis on the tasks of optical flow, depth estimation, rotational motion estimation, differential homography, and 6-DoF motion tracking, respectively.
}
\label{fig:numerical_test}
\end{figure*}

\begin{figure}[htbp]
  \centering
    \includegraphics[width=\textwidth]{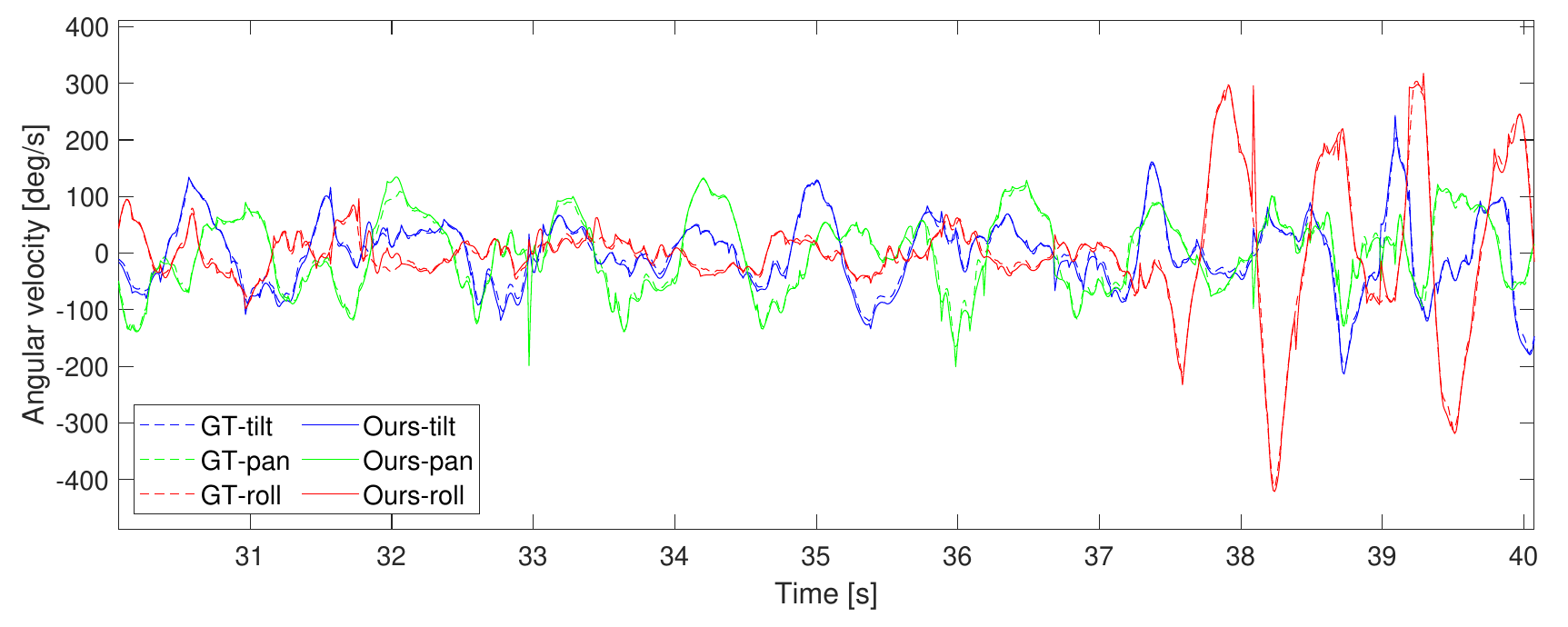}
  \caption{Results of the continuous-time angular velocity estimator on sequence \emph{dynamic\_rotation}.}
\label{fig:dynamic_rotation_results}
\end{figure}

\subsection{Numerical Stability Analysis}
\label{sec:numerical_supp}
We conduct the numerical stability analysis of the linear solver in the five problems tackled: 
1) Optical flow estimation, 
2) Depth estimation,
3) Rotational estimation, 
4) Differential homography estimation, 
and 5) 6-DoF motion tracking. 
We gradually increase the noise level from 0.01 px to 100 px on the normal flow observations, and assess how sensitive the linear solver is in each problem. 

As shown in \cref{fig:numerical_test}, for optical flow estimation, the error increases with the noise level and the linear solver can estimate optical flow under 1-pixel error in normal flow observation.
Also, the linear solver can estimate depth with high accuracy under the same noise level.
For tasks involving rotation, differential homography, and 6-DoF motion estimation, the linear solver demonstrates greater resilience, effectively functioning under more severe noise conditions.
In general, accurate results are witnessed in all five problems under a noise level of 1 pixel, thereby substantiating the robustness of our linear solver.

\subsection{Continuous-Time Nonlinear Solver}
The experiments on sequence \emph{shapes\_rotation} from the dataset~\cite{Mueggler17ijrr} have already demonstrated the superiority of the continuous-time nonlinear solver in handling aggressive motion. 
However, the relatively simple texture of this sequence raises questions about the solver's performance in more complex visual environments. 
To address this, we conduct an additional experiment on sequence \emph{dynamic\_rotation}, also from the dataset~\cite{Mueggler17ijrr}, which features a more natural and higher textured scene. 
The results, illustrated in \cref{fig:dynamic_rotation_results}, confirm the solver's capability to perform accurately in everyday-textured scenes, indicating its robustness across diverse scene complexities.

\bibliographystyle{splncs04}
\bibliography{chapters/07_references}

\end{document}